\documentclass[
sigconf = true, 
review = false, 
screen = true, 
anonymous = false, 
nonacm = true,
]{acmart}

\usepackage{graphicx}
\usepackage{mathtools}
\usepackage{xspace}
\usepackage{csquotes}
\usepackage{wasysym}
\usepackage[vlined,linesnumbered,ruled,titlenotnumbered]
{algorithm2e}
\usepackage{xcolor}
\usepackage{url}
\usepackage{tikz}
\usepackage{float}
\usepackage{booktabs}
\usepackage{makecell}
\usepackage{multicol}
\usepackage{multirow}
\usepackage{colortbl}
\usepackage{longtable}
\usepackage{array}
\usepackage{anyfontsize}
\usepackage{balance}
\usepackage{subcaption}
\usepackage{pifont}

\DeclareMathOperator*{\argmax}{arg\,max}

\newcommand{\ignore}[1]{}


\begin{document}


\author{Jakob Bossek}
\affiliation{%
  \institution{Statistics and Optimization\\Dept. of Information Systems\\University of M\"unster}
  \city{M\"unster, Germany}
  \country{}
}

\author{Aneta Neumann}
\affiliation{%
  \institution{Optimisation and Logistics\\School of Computer Science\\The University of Adelaide}
  \city{Adelaide, Australia}
  \country{}
}

\author{Frank Neumann}
\affiliation{%
  \institution{Optimisation and Logistics\\School of Computer Science\\The University of Adelaide}
  \city{Adelaide, Australia}
  \country{}
}

\title{Breeding Diverse Packings for the Knapsack Problem by Means of Diversity-Tailored Evolutionary Algorithms}

\begin{abstract}
In practise, it is often desirable to provide the decision-maker with a rich set of diverse solutions of decent quality instead of just a single solution.
In this paper we study evolutionary diversity optimization for the knapsack problem (KP). Our goal is to evolve a population of solutions that all have a profit of at least $(1-\varepsilon)\cdot OPT$, where OPT is the value of an optimal solution. Furthermore, they should differ in structure with respect to an entropy-based diversity measure. To this end we propose a simple $(\mu+1)$-EA with initial approximate solutions calculated by a well-known FPTAS for the KP. We investigate the effect of different standard mutation operators and introduce biased mutation and crossover which puts strong probability on flipping bits of low and/or high frequency within the population. An experimental study on different instances and settings shows that the proposed mutation operators in most cases perform slightly inferior in the long term, but show strong benefits if the number of function evaluations is severely limited.
\end{abstract}

\maketitle

\section{Introduction}

Creating diverse sets of high quality solutions has gained increasing interest in the evolutionary computation literature. The two prominent terms associated with this research are quality diversity (QD) algorithms and evolutionary diversity optimization (EDO). Both areas have almost developed independently but share the same goal whereas QD algorithms are mainly developed in the areas of machine learning and robotics and EDO approaches are focused on areas such as combinatorial optimization and image design.

Quality diversity algorithms are able to generate many solutions with a diverse set of behaviors by maintaining a diverse archive of high performing solutions. In~\cite{DBLP:conf/gecco/LehmanS11}, quality diversity is explicitly encouraged by rewarding diverse behavior and local competition i.e., individuals are kept that outperform those most similar in the behavior space. The main difference between the QD algorithms and standard evolutionary approaches is the definition of an archive and the selection process~\cite{DBLP:conf/gecco/LehmanS11,DBLP:journals/ec/MouretD12,DBLP:journals/corr/MouretC15}. In recent years QD approaches were successfully applied in areas such as real world design processes~\cite{DBLP:conf/ppsn/HaggAB18} and robotics~\cite{DBLP:journals/ec/MouretD12}.

Evolutionary diversity optimization was introduced in~\cite{DBLP:conf/gecco/UlrichT11} and the explicit aim is to produce a set of high quality solutions with a high diversity. In contrast to other evolutionary approaches that maintain diverse solutions in order to prevent premature convergence during the optimization, maximizing diversity allows to learn more about the optimization problem itself and provides the decision maker a diverse set of high quality solutions to choose from.

In the context of EDO, most studies have so far focused on diversity measure for creating high quality diverse sets of solutions. Such approaches map solutions to feature values and try to diversify the set of solutions with respect to the given features. Studies include basic approaches on weighted differences in feature values applied to the creation of diverse sets of TSP instances~\cite{doi:10.1162/evcoa00274} and images~\cite{DBLP:conf/gecco/AlexanderKN17}. Furthermore, indicator-based approaches have been studied which map the set of solutions to one scalar value. The first approach of this type has used the (star) discrepancy measure~\cite{DBLP:conf/gecco/NeumannGDN018} which is widely studied in the area of mathematics~\cite{JOUR:Niederreiter72,DBLP:journals/tog/DobkinEM96}. Furthermore, it has been shown how to apply performance indicators from the area of evolutionary multi-objective optimization such as the hypervolume indicator (HYP) and inverse generational distance (IGD) to EDO~\cite{DBLP:conf/gecco/NeumannG0019}. The results show that the  HYP and IGD are very well suited for computing high quality diverse sets of TSP instances and images for a wide range of features.

The investigation of EDO approaches for classical combinatorial optimization problems has only started very recently. Do et al.~\cite{DBLP:conf/gecco/DoBN020} have investigated the use of different diversity measures with respect to the edges included in a TSP tour and designed basic EDO approaches using these diversity measure to obtain diverse sets of high quality TSP tours. In the context of submodular optimization, diversifying greedy sampling approaches and their combination with EDO have been introduced in \cite{DBLP:journals/corr/abs-2010-11486}. The results in this paper show that the investigated approaches provably obtain good approximations for monotone submodular problems and achieve good diversity in practice for submodular problems such as influence maximization in social networks and maximum coverage in graphs.

In this paper, we investigate EDO for the classical knapsack problem and aim to create high quality diverse sets of solutions where the solutions differ with respect to the items that they include. Solutions are represented as binary strings. The diversity of a set of solutions is measured in terms of an entropy measure based on the fraction of number of times an item appears in a given population. The measure incentives to include components that are less frequent in the current population.
We study how different types of variation operators can be used to create high quality diverse sets of solutions.  Our goal is to create diverse sets of solutions where all solutions are a good approximation of an optimal solution.
The most basic variation operator when dealing with binary strings are mutation operators flipping a single bit or standard bit mutation where each bit is flipped in each mutation step with probability $1/n$. When designing specific operator for EDO, two conflicting goals have to be taken into account. On the one hand, it is necessary to construct solutions that meet the given quality threshold set for high quality solutions. On the other hand, the goal is to construct solutions that are different the solution contained in the current set of solutions. We introduce mutation and crossover operators that allow to diversify the set of solutions by preferring solution components that are not contained in the current population. In the case of mutation, we study biased and heavy tailed mutation operators. In the case of crossover, we propose a method that focuses on less frequent items greedily.
In our experimental study, we investigate the different operators for a wide range of knapsack instances, approximation thresholds and population sizes.
Our experimental investigations reveal that the new mutation operators significantly improve the EDO process if the computational budget is severely limited, but are slightly outperformed by standard operators in rather long runs. It also shows that crossover is beneficial in combination with biased mutation.

The paper is structured as follows. In Section~\ref{sec:knapsack_problem} we formally describe the knapsack problem in the context of EDO. In Section~\ref{sec:evolutionary_algorithm} we introduce a simple evolutionary algorithm for diversity optimization and different variation operators that favor knapsack diversity. We present our experimental investigation and discuss experimental results in Sections~\ref{sec:experimental_setup} and \ref{sec:experimental_results}. Finally, we finish with some concluding remarks.

\section{The Knapsack Problem and EDO}
\label{sec:knapsack_problem}

In the following, we use the standard notation $[n]=\{1,\ldots,n\}$ to express the set of the first $n$ positive integers.
The problem studied is the classical \emph{zero-one knapsack problem} (KP). We are given a knapsack with finite integer capacity $W>0$ and a set of $n$ items. Each item is associated with a positive integer weight $w_i$ and integer profit/value $v_i$. Each subset $s \subset [n]$ is called a \emph{solution/packing}. We identify a packing by means of a binary vector $x \in \{0,1\}^n$ where $x_i=1$ means that the $i$-\emph{th} item is packed (it is \emph{active}) while $x_i=0$ indicates that it is not packed (\emph{inactive}). We write
\begin{align*}
    w(x) = \sum_{i=1}^{n} x_iw_i \text{ and } v(x) = \sum_{i=1}^{n} x_iv_i
\end{align*}
for the total weight and value respectively. A solution is \emph{feasible} if its total weight does not exceed the capacity. Let
\begin{align*}
    \mathcal{S} = \{x \mid x \in \{0,1\}^n \wedge w(x) \leq W\}
\end{align*}
be the set of feasible solutions. The goal in the optimization version of the problem is to find a feasible solution $x^{*} \in \mathcal{S}$ with maximum profit $v(x^*)$.

The knapsack problem is a well-studied classical NP-hard optimization problem. However, it is weakly NP-hard in the sense that it admits fully polynomial time approximation schemes (FPTAS), i.e., algorithms that for an arbitrary input instance and a parameter $0 < \varepsilon < 1$ output a solution $x \in \mathcal{S}$ with
\begin{align*}
v(x) \geq (1 - \varepsilon) \cdot OPT
\end{align*}
where $OPT$ is the unknown optimal value. The classical textbook FPTAS runs in $O(n^3/\varepsilon)$ which is polynomial both in $n$ and $1/\varepsilon$~\cite{Vazirani2010ApproxAlgorithms}.

Evolutionary diversity optimization~(EDO) aims to evolve a set of minimum-quality solutions that differ in structure. More precisely, in the context of the KP, given a quality threshold $v_{min}$ the goal is to evolve a population $P$ of size $\mu=|P|$ of packings such that (1) for all $x \in P$ it holds that $w(x) \leq W$ and $v(x) \geq v_{min}$ and (2) $P$ is of maximum diversity with respect to a diversity measure. Denote by
\begin{align*}
h(i) := |\{x \in P \,|\, x_i=1\}|
\end{align*}
the absolute frequency of item $i$ in the population. Then $f(i) := \frac{h(i)}{\mu}$ for $1 \leq i \leq n$ is the share of individuals in the population that contain item $i$. We aim to maximize the \emph{entropy measure}
\begin{align*}
H(P) := -\sum_{i=1}^{n} f(i) \cdot \log f(i).
\end{align*}
Note that the contribution of an item that is not included in neither packing is zero and so is the contribution of an item that is packed in all solutions. The entropy measure guides the EA to include less frequent items and drop very frequent ones.

\section{Evolutionary Approach}
\label{sec:evolutionary_algorithm}

By running the classical textbook FPTAS for the knapsack problem~\cite{Vazirani2010ApproxAlgorithms} with approximation parameter $\epsilon/2$ we obtain a packing $x'$ which is at most a factor of $(1-\varepsilon/2)$ less than the value of an optimal solution denoted as OPT. This provides a good starting point to seed the population of an EDO-focused evolutionary algorithm and use $v(x')$ as the minimum quality threshold. However, in general we do not know where exactly in $[(1-\varepsilon/2)\cdot OPT, OPT]$ the solution $v(x')$ is located; it may be very close to $OPT$. This is a bad setup for EDO as there may be a unique optimum or the set of global optima may be very small leaving literally no chance to the EA to evolve a diverse population in a meaningful way. Therefore, the diversifying EA requires all solutions $x$ in the population $P$ to admit
\begin{align*}
v(x) \geq (1-\varepsilon/2) \cdot v(x').
\end{align*}
This means that we allow it to deviate from the approximate solution quality by another multiplicative factor of $(1-\varepsilon/2)$. The choice $\varepsilon/2$ leads to the following quality guarantee of all solutions in the final population:
\begin{align*}
v(x)
& \geq (1-\varepsilon/2)\cdot v(x') \\
& \geq (1-\varepsilon/2)\cdot(1-\varepsilon/2)\cdot OPT\\
& = (1-\varepsilon/2)^2 \cdot OPT \\
& \geq (1-\varepsilon) \cdot OPT.
\end{align*}
Here, the last transformation is due to Bernoulli's inequality $(1+x)^n \geq 1+nx$ for $x \geq -1$ and the fact that $-\varepsilon/2 \geq 1$ for all reasonable values of $\varepsilon$.

\begin{algorithm}[t]
\SetKwInOut{Input}{Input}
\Input{Initial solution $x'$, $0 < \varepsilon < 1$, population size $\mu$, crossover probability $p_c \in [0,1]$, repair (on/off)}
Set $v_{min}$ to $(1-\varepsilon/2) \cdot v(x')$\;
Initialize $P$ with $\mu$ copies of $x'$\;
\While{termination condition not met}{
    With probability $p_c$ generate $x$ from two random individuals $x^1, x^2$ from $P$ by crossover (see Algorithm~\ref{alg:crossover}), otherwise sample $x$ randomly from $P$\;
    Modify $x$ by mutation\;
    \If{repair is on}{
        Call Algorithm~\ref{alg:repair} passing $x$, $v_{min}$ and $W$\;
    }
    \If{$w(x) \leq W \land v(x) \geq v_{min}$}{
        Add $x$ to $P$\;
    }
    \If{$|P| = \mu + 1$}{
        Remove $y \in P$ with $y = \argmax_{y' \in P} H(P \setminus \{y'\})$\;
    }
}
\Return{$P$}\;
\caption{$(\mu+1)$-EA for EDO.}
\label{alg:diversity_maximizing_ea}
\end{algorithm}

We now introduce a simple evolutionary algorithm for diversity optimization. The algorithm is a classical $(\mu+1)$-EA outlined in Algorithm~\ref{alg:diversity_maximizing_ea}. The algorithm's initial population is seeded with $\mu$ copies of a $(1-\varepsilon/2)$-approximate solution $x'$ calculated by the FPTAS as discussed above. We set $v_{min} = (1-\varepsilon/2) \cdot v(x')$ (line~2) and in the course of optimization reject solutions that violate the capacity constraint $W$ or do not meet the minimum quality $v_{min}$. This way the algorithm ensures all solutions to be at most up to a multiplicative factor of $(1-\varepsilon)$ away from the optimum. Note that after initialization all individuals are feasible.
In one iteration of the EA, exactly one individual $x$ is produced either by crossover applied to two random individuals $x^1, x^2 \in P$ with a probability $p_c \in [0, 1]$ or by copying a random individual from $P$ with inverse probability $(1-p_c)$. Next $x$ undergoes mutation followed by a repair procedure which tries to re-establish feasibility of $x$ if it is infeasible. Afterwards, $x$ is added to the population if neither the knapsack capacity nor the least-quality is violated. Finally, to ensure a constant population size $\mu$, the algorithm drops $y = \argmax_{y' \in P} H(P \setminus \{y'\})$, i.e., the individual that leads to the maximum population entropy when dropped in the current iteration.\footnote{The survival selection is the computationally most demanding operation in this algorithm with a runtime of $O(n\mu)$ per iterations. It can be replaced by a simplified theme where the added packing $x$ competes against (one of ) its parent(s) only. However, in this paper we aim for the maximal possible increase in each iteration.}

We study Algorithm~\ref{alg:diversity_maximizing_ea} with different variations operators discussed in detail in the following.

\subsection{Mutation operators}

Mutation is key to exploration of the search space. We study five different mutation operators in total, three of which are classical operators from the literature and the other two being tailored towards diversity. The classical operators are:

\paragraph{Standard bit-flip (BF)} This is the unbiased baseline. Each bit is flipped independently with mutation probability $p=1/n$. Note that in expectation exactly one bit is flipped and there is a quite high probability of $(1-1/n)^n \approx e^{-1}$ to leave all bits untouched. For EDO it seems advantageous to flip more than one bit with higher probability, e.g., in order to activate a previously inactive item and deactivate an active item. The probability to flip multiple bits is highly increased for the following two mutation operators.

\paragraph{Poisson bit-flip (PBF)} A value $k = 1 \,+\, \text{Pois}(1)$ is sampled first and bits at $k$ randomly selected positions are flipped. Here, $\text{Pois}(1)$ describes a random integer value sampled from a Poisson-distribution with rate parameter $\lambda=1$. Since the expected value of a Poisson-distribution corresponds to the reciprocal of its rate, in expectation this operator flips two bits and in any case flips at least one bit.

\paragraph{Heavy-tailed bit-flip (HTBF)} This mutation operator was proposed in Friedrich~et~al.~\cite{Friedrich2018HeavyTailed} and was shown to perform excellently in theory and experimental evaluations. It is particularly suited to overcome fitness plateaus. The operator works as follows: the mutation rate is sampled randomly in each mutation following a power-law distribution with exponent $\beta>1$. I.e., sample $\theta \in [1, \ldots, n/2]$ from the $D_{n/2}^{\beta}$-distribution such that
\begin{align*}
\text{Prob}(\text{sample } \theta) = \left(C_{n/2}^{\beta}\right)^{-1}\theta^{-\beta}
\end{align*}
for all $\theta \in [1, \ldots, n/2]$ where $C_{n/2}^{\beta}=\sum_{i=1}^{n/2} i^{-\beta}$. Given $\theta$, each bit is flipped independently with probability $p=\theta/n$.

We expect PBF and HTBF to outperform BF in EDO. However, although PBF and HTBF may flip multiple bits and thus increase entropy faster, they may flip the wrong bits, e.g., mostly inactive or mostly active bits which likely leads to mutants being infeasible violating one of the two constraints. We therefore propose two mutation operators that take into account the current item frequencies $h(i), 1 \leq i \leq n$ and work with a strong bias towards more diverse packings potentially.

\paragraph{EDO biased bit-flip~1 (EDO-BBF1)} Here we use the idea of asymmetric mutation probabilities~\cite{DBLP:journals/ec/DoerrHN07}. We put higher probability on activating items that are inactive, but with low frequency. Analogously, we increase the probability to deactivate frequent items. To be more precise the probability $p_i$ to flip the $i$th item is given by
\begin{align*}
p_i =
\begin{cases}
\frac{\mu-h_i}{2n} & \text{ if } x_i=0 \text{ and } h_i \leq \frac{\mu}{2} \\
\frac{h_i}{2n} & \text{ if } x_i=1 \text{ and } h_i > \frac{\mu}{2}.
\end{cases}
\end{align*}
For instance, if $\mu=50$ and $n=100$, after the initialization of Algorithm~\ref{alg:diversity_maximizing_ea}, we have $h(i) = \mu$ for items that are packed in the FPTAS-solution and $h(i)=0$ otherwise. Therefore, both active and inactive items would have a very high probability of $1/4$ to flip in the first iteration. This operator is rather extreme and due to the described effect we expect it to achieve reasonable diversity values in short time. However, note that strong bias may also have the reverse effect if the subset of feasible solutions is small in size and flipping many bits is likely to generate infeasible solutions.

\paragraph{EDO biased bit-flip~2 (EDO-BBF2)} This operator aims to balance bit-flips of active and inactive items, but does not rely on the frequencies and is less \enquote{extreme} than EDO-BBF1. To this end let $X_j = \{i \in [n] \, | \, x_i = j\}, j=0, 1$ be the set of (in)active items in $x$. The algorithm samples two numbers $k_1,k_2$ from a $1 + \text{Pois}(1)$. Next, it samples $\min\{k_0, |X_0|\}$ bit positions from $X_0$ and $\min\{k_1, |X_1|\}$ bit positions from $X_1$ for flipping. The minimum is necessary as in certain situations the number of zero- or one-bits may be lower than the sampled number. Note that $k_1+k_2 \geq 2$, i.e., at least two bits are flipped.

\subsection{EDO-focused repair operator}

\begin{algorithm}[t]
\SetKwInOut{Input}{Input}
\Input{Solution $x$, minimal quality $v_{min}$, capacity $W$}
\If{$w(x) > W$}{
    Drop items from $x$ in decreasing order of frequency $h(i)$\;
}
\If{$v(x) < v_{min}$}{
    Add items to $x$ in increasing order of frequency $h(i)$\;
}
\Return{$x$}\;
\caption{EDO-focused repair operator.}
\label{alg:repair}
\end{algorithm}
\begin{algorithm}[t]
\SetKwInOut{Input}{Input}
\Input{Parents $x^1, x^2$, minimal quality $v_{min}$}
Set $x_i = 1$ if $x^1_i = 1 \land x^2_i = 1$ and $x_i=0$ otherwise\;
\If{$v(x) < v_{min}$}{
    Add items to $x$ in increasing order of frequency $h(i)$\;
}
\Return{$x$}\;
\caption{EDO-focused crossover.}
\label{alg:crossover}
\end{algorithm}

Mutated individuals may be infeasible if either the allowed capacity limit $W$ is exceeded or the minimum quality is not achieved. It is plausible to assume that this effect appears more often if $\varepsilon$ is low and in consequence there is little flexibility to deviate from the initial solutions' quality. To account for this problem we propose a simple repair operator outlined in Algorithm~\ref{alg:repair}. The repair operator first sorts the items in increasing order of their frequencies $h(i), i \in [n]$. It then first traverses the items in reverse order of item frequency if the knapsack capacity is violated. Inactive items $i \in [n]$, i.e. where $x_i=0$, are ignored, while active items are flipped until $w(x) \leq W$. In a second phase, given the quality threshold is violated, the items are traversed in increasing order of frequency, thus activating the least frequent non-active item and so on until the quality adheres to the threshold. Certainly, this approach cannot guarantee to be successful. E.g, the bias towards item frequency may lead to effects where the first phase manages to fix the capacity overshooting, but in the subsequent phase may add a heavy item that does not occur often in the population leading to a repeated violation of the capacity constraint. This issue could be partly fixed by checking the other constraint before activating (in phase one) or deactivating (in phase two). However, preliminary experiments showed no noticeable difference in the behavior. It should be noted that the repair operator runtime is dominated by sorting the items. We can leverage the fact that item frequencies are integer values in the interval $[0, \ldots, \mu]$ and adopt a linear-time sorting algorithm like counting-sort~\cite{CLRS2009}. Thus, the operator can be implemented efficiently in time $O(n + \mu) = O(\max\{n, \mu\})$ which is $O(n)$ if $\mu = O(n)$.

\subsection{EDO-focused crossover}

Our crossover operator expects two parents $x^1, x^2$ and produces one child $x$ (see Algorithm~\ref{alg:crossover}). It first transfers all items common to both parents, i.e. set $x_i=1$ if both $x^1_i = 1$ and $y^2_i = 1$. The rationale behind this is to keep important items that may be necessary to include into every single feasible solution due to their high efficiency (high profit and low weight). After this step the number of active items in $x$ is less than or equal to the number of active items in $x^1$ or $x^2$ respectively. Moreover, $w(x) \leq \min\{w(x^1), w(x^2)\}$ and hence the capacity limit is certainly not violated. In a second step the algorithm uses phase two of the repair operator (cf.~Algorithm~\ref{alg:repair}) to fill the knapsack with further unpacked items in increasing order of frequency in the population. The operator also has a worst case complexity of $O(n + \mu)$.

\begin{table*}[htbp]

\caption{\label{tab:}Mean (\text{mean}) and standard deviation (\text{std}) of mean entropy. Highest mean values are highlighted in \colorbox{gray!20}{\textbf{bold face}}. Algorithms are numbered (see second row). In the stat-column a number $X$ means that the algorithm in the respective column has significantly higher entropy than $X$ while $X-Y$ means that it is superior to all algorithms $X \ldots Y$.}
\label{tab:entropy_mutation_only_with_repair}
\renewcommand{\arraystretch}{0.58}
\renewcommand{\tabcolsep}{7pt}
\centering
\begin{scriptsize}\begin{tabular}[t]{crrrrrrrrrrrrrrrrrr}
\toprule
\multicolumn{1}{c}{\textbf{ }} & \multicolumn{1}{c}{\textbf{ }} & \multicolumn{1}{c}{\textbf{ }} & \multicolumn{1}{c}{\textbf{ }} & \multicolumn{9}{c}{\textbf{Standard}} & \multicolumn{6}{c}{\textbf{Biased}} \\
\cmidrule(l{3pt}r{3pt}){5-13} \cmidrule(l{3pt}r{3pt}){14-19}
\multicolumn{1}{c}{\textbf{ }} & \multicolumn{1}{c}{\textbf{ }} & \multicolumn{1}{c}{\textbf{ }} & \multicolumn{1}{c}{\textbf{ }} & \multicolumn{3}{c}{\textbf{BF (1)}} & \multicolumn{3}{c}{\textbf{PBF (2)}} & \multicolumn{3}{c}{\textbf{HTBF (3)}} & \multicolumn{3}{c}{\textbf{EDO-BBF1 (4)}} & \multicolumn{3}{c}{\textbf{EDO-BBF2 (5)}} \\
\cmidrule(l{3pt}r{3pt}){5-7} \cmidrule(l{3pt}r{3pt}){8-10} \cmidrule(l{3pt}r{3pt}){11-13} \cmidrule(l{3pt}r{3pt}){14-16} \cmidrule(l{3pt}r{3pt}){17-19}
 & $D$ & $\mu$ & $\varepsilon$ & \textbf{mean} & \textbf{std} & \textbf{stat} & \textbf{mean} & \textbf{std} & \textbf{stat} & \textbf{mean} & \textbf{std} & \textbf{stat} & \textbf{mean} & \textbf{std} & \textbf{stat} & \textbf{mean} & \textbf{std} & \textbf{stat}\\
\midrule
 &  &  & 0.1 & 36.74 & 0.02 & 4-5 & \cellcolor{gray!20}{\textbf{36.77}} & 0.00 & 1,3-5 & 36.76 & 0.01 & 1,4-5 & 36.36 & 0.11 &  & 36.40 & 0.01 & \\

 &  &  & 0.5 & 36.75 & 0.01 & 4-5 & \cellcolor{gray!20}{\textbf{36.77}} & 0.00 & 1,3-5 & 36.76 & 0.00 & 1,4-5 & 36.38 & 0.10 &  & 36.40 & 0.01 & \\

 &  & \multirow{-3}{*}{\raggedleft\arraybackslash 25} & 0.9 & 36.74 & 0.02 & 4-5 & \cellcolor{gray!20}{\textbf{36.77}} & 0.00 & 1,3-5 & \cellcolor{gray!20}{\textbf{36.77}} & 0.00 & 1,4-5 & 36.42 & 0.13 &  & 36.40 & 0.01 & \\

\cmidrule{3-19}
 &  &  & 0.1 & \cellcolor{gray!20}{\textbf{36.79}} & 0.00 & 4-5 & \cellcolor{gray!20}{\textbf{36.79}} & 0.00 & 4-5 & \cellcolor{gray!20}{\textbf{36.79}} & 0.00 & 4-5 & 36.58 & 0.02 &  & 36.71 & 0.04 & 4\\

 &  &  & 0.5 & \cellcolor{gray!20}{\textbf{36.79}} & 0.00 & 4-5 & \cellcolor{gray!20}{\textbf{36.79}} & 0.00 & 4-5 & \cellcolor{gray!20}{\textbf{36.79}} & 0.00 & 4-5 & 36.61 & 0.02 &  & 36.73 & 0.03 & 4\\

 & \multirow{-8}{*}{\raggedleft\arraybackslash 2} & \multirow{-3}{*}{\raggedleft\arraybackslash 100} & 0.9 & \cellcolor{gray!20}{\textbf{36.79}} & 0.00 & 4-5 & \cellcolor{gray!20}{\textbf{36.79}} & 0.00 & 4-5 & \cellcolor{gray!20}{\textbf{36.79}} & 0.00 & 4-5 & 36.60 & 0.02 &  & 36.72 & 0.03 & 4\\

\cmidrule{2-19}
 &  &  & 0.1 & 36.06 & 0.02 & 4-5 & \cellcolor{gray!20}{\textbf{36.10}} & 0.01 & 1,3-5 & 36.09 & 0.01 & 1,4-5 & 35.85 & 0.03 &  & 35.91 & 0.04 & 4\\

 &  &  & 0.5 & 36.75 & 0.01 & 4-5 & \cellcolor{gray!20}{\textbf{36.77}} & 0.00 & 1,3-5 & 36.76 & 0.00 & 1,4-5 & 36.11 & 0.08 &  & 36.21 & 0.13 & 4\\

 &  & \multirow{-3}{*}{\raggedleft\arraybackslash 25} & 0.9 & 36.75 & 0.01 & 4-5 & \cellcolor{gray!20}{\textbf{36.77}} & 0.00 & 1,3-5 & 36.76 & 0.01 & 1,4-5 & 36.08 & 0.08 &  & 36.28 & 0.12 & 4\\

\cmidrule{3-19}
 &  &  & 0.1 & \cellcolor{gray!20}{\textbf{36.17}} & 0.00 & 4-5 & \cellcolor{gray!20}{\textbf{36.17}} & 0.00 & 1,4-5 & \cellcolor{gray!20}{\textbf{36.17}} & 0.00 & 1,4-5 & 35.80 & 0.02 &  & 36.15 & 0.01 & 4\\

 &  &  & 0.5 & \cellcolor{gray!20}{\textbf{36.79}} & 0.00 & 4-5 & \cellcolor{gray!20}{\textbf{36.79}} & 0.00 & 4-5 & \cellcolor{gray!20}{\textbf{36.79}} & 0.00 & 4-5 & 35.87 & 0.02 &  & 36.66 & 0.06 & 4\\

 & \multirow{-8}{*}{\raggedleft\arraybackslash 5} & \multirow{-3}{*}{\raggedleft\arraybackslash 100} & 0.9 & \cellcolor{gray!20}{\textbf{36.79}} & 0.00 & 4-5 & \cellcolor{gray!20}{\textbf{36.79}} & 0.00 & 4-5 & \cellcolor{gray!20}{\textbf{36.79}} & 0.00 & 4-5 & 35.86 & 0.04 &  & 36.68 & 0.05 & 4\\

\cmidrule{2-19}
 &  &  & 0.1 & \cellcolor{gray!20}{\textbf{0.00}} & 0.00 &  & \cellcolor{gray!20}{\textbf{0.00}} & 0.00 &  & \cellcolor{gray!20}{\textbf{0.00}} & 0.00 &  & \cellcolor{gray!20}{\textbf{0.00}} & 0.00 &  & \cellcolor{gray!20}{\textbf{0.00}} & 0.00 & \\

 &  &  & 0.5 & 35.85 & 0.02 & 4-5 & \cellcolor{gray!20}{\textbf{35.88}} & 0.01 & 1,3-5 & 35.86 & 0.01 & 4-5 & 35.68 & 0.05 &  & 35.76 & 0.02 & 4\\

 &  & \multirow{-3}{*}{\raggedleft\arraybackslash 25} & 0.9 & \cellcolor{gray!20}{\textbf{36.77}} & 0.01 & 3-5 & \cellcolor{gray!20}{\textbf{36.77}} & 0.00 & 1,3-5 & 36.76 & 0.00 & 4-5 & 36.17 & 0.10 &  & 36.23 & 0.15 & \\

\cmidrule{3-19}
 &  &  & 0.1 & \cellcolor{gray!20}{\textbf{0.00}} & 0.00 &  & \cellcolor{gray!20}{\textbf{0.00}} & 0.00 &  & \cellcolor{gray!20}{\textbf{0.00}} & 0.00 &  & \cellcolor{gray!20}{\textbf{0.00}} & 0.00 &  & \cellcolor{gray!20}{\textbf{0.00}} & 0.00 & \\

 &  &  & 0.5 & \cellcolor{gray!20}{\textbf{35.95}} & 0.00 & 4 & \cellcolor{gray!20}{\textbf{35.95}} & 0.00 & 1,3-5 & \cellcolor{gray!20}{\textbf{35.95}} & 0.00 & 1,4-5 & 35.58 & 0.02 &  & 35.94 & 0.01 & 4\\

\multirow{-20}{*}{\centering\arraybackslash \rotatebox[origin=c]{90}{\hskip25pt scorr}} & \multirow{-8}{*}{\raggedleft\arraybackslash 10} & \multirow{-3}{*}{\raggedleft\arraybackslash 100} & 0.9 & \cellcolor{gray!20}{\textbf{36.79}} & 0.00 & 4-5 & \cellcolor{gray!20}{\textbf{36.79}} & 0.00 & 4-5 & \cellcolor{gray!20}{\textbf{36.79}} & 0.00 & 4-5 & 36.63 & 0.06 &  & 36.70 & 0.04 & 4\\
\cmidrule{1-19}
\cmidrule{3-19}
 &  &  & 0.1 & 36.26 & 0.01 & 4-5 & \cellcolor{gray!20}{\textbf{36.29}} & 0.01 & 1,3-5 & 36.28 & 0.01 & 1,4-5 & 36.08 & 0.07 &  & 36.14 & 0.04 & 4\\

 &  &  & 0.5 & 36.75 & 0.01 & 4-5 & \cellcolor{gray!20}{\textbf{36.77}} & 0.00 & 1,3-5 & 36.76 & 0.00 & 1,4-5 & 36.40 & 0.11 &  & 36.73 & 0.01 & 4\\

 &  & \multirow{-3}{*}{\raggedleft\arraybackslash 25} & 0.9 & 36.74 & 0.01 & 4 & \cellcolor{gray!20}{\textbf{36.77}} & 0.00 & 1,3-5 & 36.76 & 0.00 & 1,4-5 & 36.39 & 0.04 &  & 36.74 & 0.01 & 4\\

\cmidrule{3-19}
 &  &  & 0.1 & \cellcolor{gray!20}{\textbf{36.34}} & 0.00 & 4-5 & \cellcolor{gray!20}{\textbf{36.34}} & 0.00 & 1,3-5 & \cellcolor{gray!20}{\textbf{36.34}} & 0.00 & 4-5 & 36.00 & 0.04 &  & 36.33 & 0.00 & 4\\

 &  &  & 0.5 & \cellcolor{gray!20}{\textbf{36.79}} & 0.00 & 4-5 & \cellcolor{gray!20}{\textbf{36.79}} & 0.00 & 4-5 & \cellcolor{gray!20}{\textbf{36.79}} & 0.00 & 4-5 & 36.61 & 0.02 &  & \cellcolor{gray!20}{\textbf{36.79}} & 0.00 & 4\\

 & \multirow{-8}{*}{\raggedleft\arraybackslash 2} & \multirow{-3}{*}{\raggedleft\arraybackslash 100} & 0.9 & \cellcolor{gray!20}{\textbf{36.79}} & 0.00 & 4-5 & \cellcolor{gray!20}{\textbf{36.79}} & 0.00 & 4-5 & \cellcolor{gray!20}{\textbf{36.79}} & 0.00 & 4-5 & 36.63 & 0.02 &  & \cellcolor{gray!20}{\textbf{36.79}} & 0.00 & 4\\

\cmidrule{2-19}
 &  &  & 0.1 & 25.42 & 0.16 & 4 & \cellcolor{gray!20}{\textbf{25.65}} & 0.13 & 1,3-4 & 25.48 & 0.12 & 4 & 24.23 & 0.20 &  & 25.63 & 0.09 & 1,3-4\\

 &  &  & 0.5 & 36.68 & 0.01 & 4-5 & \cellcolor{gray!20}{\textbf{36.70}} & 0.00 & 1,3-5 & 36.69 & 0.01 & 1,4-5 & 36.02 & 0.09 &  & 36.10 & 0.09 & 4\\

 &  & \multirow{-3}{*}{\raggedleft\arraybackslash 25} & 0.9 & 36.75 & 0.01 & 4-5 & \cellcolor{gray!20}{\textbf{36.77}} & 0.00 & 1,3-5 & 36.76 & 0.00 & 1,4-5 & 36.15 & 0.03 &  & 36.15 & 0.08 & \\

\cmidrule{3-19}
 &  &  & 0.1 & 26.03 & 0.06 & 4 & \cellcolor{gray!20}{\textbf{26.14}} & 0.03 & 1,3-5 & 26.09 & 0.05 & 1,4 & 0.00 & 0.00 &  & 26.09 & 0.03 & 1,4\\

 &  &  & 0.5 & \cellcolor{gray!20}{\textbf{36.75}} & 0.00 & 3-5 & \cellcolor{gray!20}{\textbf{36.75}} & 0.00 & 1,3-5 & \cellcolor{gray!20}{\textbf{36.75}} & 0.00 & 4-5 & 35.56 & 0.03 &  & 36.63 & 0.02 & 4\\

 & \multirow{-8}{*}{\raggedleft\arraybackslash 5} & \multirow{-3}{*}{\raggedleft\arraybackslash 100} & 0.9 & \cellcolor{gray!20}{\textbf{36.79}} & 0.00 & 4-5 & \cellcolor{gray!20}{\textbf{36.79}} & 0.00 & 4-5 & \cellcolor{gray!20}{\textbf{36.79}} & 0.00 & 4-5 & 35.60 & 0.05 &  & 36.65 & 0.06 & 4\\

\cmidrule{2-19}
 &  &  & 0.1 & \cellcolor{gray!20}{\textbf{0.00}} & 0.00 &  & \cellcolor{gray!20}{\textbf{0.00}} & 0.00 &  & \cellcolor{gray!20}{\textbf{0.00}} & 0.00 &  & \cellcolor{gray!20}{\textbf{0.00}} & 0.00 &  & \cellcolor{gray!20}{\textbf{0.00}} & 0.00 & \\

 &  &  & 0.5 & 35.15 & 0.01 & 4-5 & \cellcolor{gray!20}{\textbf{35.17}} & 0.01 & 1,3-5 & 35.15 & 0.01 & 4-5 & 35.07 & 0.01 &  & 35.10 & 0.02 & 4\\

 &  & \multirow{-3}{*}{\raggedleft\arraybackslash 25} & 0.9 & 36.76 & 0.01 & 4-5 & \cellcolor{gray!20}{\textbf{36.77}} & 0.00 & 1,3-5 & 36.76 & 0.00 & 4-5 & 36.15 & 0.14 &  & 36.16 & 0.08 & \\

\cmidrule{3-19}
 &  &  & 0.1 & \cellcolor{gray!20}{\textbf{0.00}} & 0.00 &  & \cellcolor{gray!20}{\textbf{0.00}} & 0.00 &  & \cellcolor{gray!20}{\textbf{0.00}} & 0.00 &  & \cellcolor{gray!20}{\textbf{0.00}} & 0.00 &  & \cellcolor{gray!20}{\textbf{0.00}} & 0.00 & \\

 &  &  & 0.5 & 35.22 & 0.00 & 4-5 & \cellcolor{gray!20}{\textbf{35.23}} & 0.00 & 1,3-5 & 35.22 & 0.00 & 4-5 & 35.00 & 0.02 &  & 35.21 & 0.00 & 4\\

\multirow{-20}{*}{\centering\arraybackslash \rotatebox[origin=c]{90}{\hskip25pt uncorr}} & \multirow{-8}{*}{\raggedleft\arraybackslash 10} & \multirow{-3}{*}{\raggedleft\arraybackslash 100} & 0.9 & \cellcolor{gray!20}{\textbf{36.79}} & 0.00 & 4-5 & \cellcolor{gray!20}{\textbf{36.79}} & 0.00 & 4-5 & \cellcolor{gray!20}{\textbf{36.79}} & 0.00 & 4-5 & 36.60 & 0.04 &  & 36.65 & 0.06 & \\
\cmidrule{1-19}
\cmidrule{3-19}
 &  &  & 0.1 & 36.76 & 0.01 & 4-5 & \cellcolor{gray!20}{\textbf{36.77}} & 0.01 & 1,3-5 & 36.76 & 0.01 & 4-5 & 36.57 & 0.07 & 5 & 36.49 & 0.03 & \\

 &  &  & 0.5 & 36.75 & 0.01 & 4-5 & \cellcolor{gray!20}{\textbf{36.77}} & 0.00 & 1,3-5 & 36.76 & 0.00 & 1,4-5 & 36.50 & 0.07 & 5 & 33.74 & 0.27 & \\

 &  & \multirow{-3}{*}{\raggedleft\arraybackslash 25} & 0.9 & 36.75 & 0.02 & 4-5 & \cellcolor{gray!20}{\textbf{36.77}} & 0.00 & 1,3-5 & \cellcolor{gray!20}{\textbf{36.77}} & 0.00 & 1,4-5 & 36.50 & 0.08 & 5 & 30.75 & 0.00 & \\

\cmidrule{3-19}
 &  &  & 0.1 & \cellcolor{gray!20}{\textbf{36.79}} & 0.00 & 4-5 & \cellcolor{gray!20}{\textbf{36.79}} & 0.00 & 4-5 & \cellcolor{gray!20}{\textbf{36.79}} & 0.00 & 4-5 & 36.68 & 0.01 &  & 36.75 & 0.01 & 4\\

 &  &  & 0.5 & \cellcolor{gray!20}{\textbf{36.79}} & 0.00 & 4-5 & \cellcolor{gray!20}{\textbf{36.79}} & 0.00 & 4-5 & \cellcolor{gray!20}{\textbf{36.79}} & 0.00 & 4-5 & 36.68 & 0.01 & 5 & 34.75 & 0.35 & \\

 & \multirow{-8}{*}{\raggedleft\arraybackslash 2} & \multirow{-3}{*}{\raggedleft\arraybackslash 100} & 0.9 & \cellcolor{gray!20}{\textbf{36.79}} & 0.00 & 4-5 & \cellcolor{gray!20}{\textbf{36.79}} & 0.00 & 4-5 & \cellcolor{gray!20}{\textbf{36.79}} & 0.00 & 4-5 & 36.69 & 0.01 & 5 & 30.86 & 0.00 & \\

\cmidrule{2-19}
 &  &  & 0.1 & 29.60 & 0.06 & 4-5 & \cellcolor{gray!20}{\textbf{29.68}} & 0.07 & 1,4-5 & 29.67 & 0.04 & 1,4-5 & 29.08 & 0.14 &  & 29.34 & 0.08 & 4\\

 &  &  & 0.5 & 36.73 & 0.01 & 4-5 & \cellcolor{gray!20}{\textbf{36.76}} & 0.01 & 1,3-5 & 36.75 & 0.00 & 1,4-5 & 35.97 & 0.10 & 5 & 35.89 & 0.01 & \\

 &  & \multirow{-3}{*}{\raggedleft\arraybackslash 25} & 0.9 & 36.74 & 0.01 & 4-5 & \cellcolor{gray!20}{\textbf{36.77}} & 0.00 & 1,3-5 & 36.76 & 0.00 & 1,4-5 & 36.06 & 0.13 & 5 & 35.89 & 0.00 & \\

\cmidrule{3-19}
 &  &  & 0.1 & 29.93 & 0.02 & 4-5 & \cellcolor{gray!20}{\textbf{29.96}} & 0.01 & 1,3-5 & 29.95 & 0.02 & 4-5 & 0.00 & 0.00 &  & 29.78 & 0.02 & 4\\

 &  &  & 0.5 & \cellcolor{gray!20}{\textbf{36.79}} & 0.00 & 4-5 & \cellcolor{gray!20}{\textbf{36.79}} & 0.00 & 4-5 & \cellcolor{gray!20}{\textbf{36.79}} & 0.00 & 4-5 & 35.39 & 0.08 &  & 35.93 & 0.00 & 4\\

 & \multirow{-8}{*}{\raggedleft\arraybackslash 5} & \multirow{-3}{*}{\raggedleft\arraybackslash 100} & 0.9 & \cellcolor{gray!20}{\textbf{36.79}} & 0.00 & 4-5 & \cellcolor{gray!20}{\textbf{36.79}} & 0.00 & 4-5 & \cellcolor{gray!20}{\textbf{36.79}} & 0.00 & 4-5 & 35.41 & 0.04 &  & 35.93 & 0.00 & 4\\

\cmidrule{2-19}
 &  &  & 0.1 & \cellcolor{gray!20}{\textbf{0.00}} & 0.00 &  & \cellcolor{gray!20}{\textbf{0.00}} & 0.00 &  & \cellcolor{gray!20}{\textbf{0.00}} & 0.00 &  & \cellcolor{gray!20}{\textbf{0.00}} & 0.00 &  & \cellcolor{gray!20}{\textbf{0.00}} & 0.00 & \\

 &  &  & 0.5 & 35.20 & 0.02 & 4-5 & \cellcolor{gray!20}{\textbf{35.23}} & 0.01 & 1,3-5 & 35.21 & 0.02 & 1,4-5 & 35.10 & 0.03 & 5 & 34.86 & 0.01 & \\

 &  & \multirow{-3}{*}{\raggedleft\arraybackslash 25} & 0.9 & 36.76 & 0.01 & 4-5 & \cellcolor{gray!20}{\textbf{36.77}} & 0.00 & 1,3-5 & 36.76 & 0.00 & 4-5 & 36.06 & 0.08 & 5 & 34.90 & 0.02 & \\

\cmidrule{3-19}
 &  &  & 0.1 & \cellcolor{gray!20}{\textbf{0.00}} & 0.00 &  & \cellcolor{gray!20}{\textbf{0.00}} & 0.00 &  & \cellcolor{gray!20}{\textbf{0.00}} & 0.00 &  & \cellcolor{gray!20}{\textbf{0.00}} & 0.00 &  & \cellcolor{gray!20}{\textbf{0.00}} & 0.00 & \\

 &  &  & 0.5 & \cellcolor{gray!20}{\textbf{35.30}} & 0.01 & 4-5 & \cellcolor{gray!20}{\textbf{35.30}} & 0.00 & 3-5 & \cellcolor{gray!20}{\textbf{35.30}} & 0.00 & 4-5 & 34.93 & 0.01 &  & 34.94 & 0.00 & 4\\

\multirow{-20}{*}{\centering\arraybackslash \rotatebox[origin=c]{90}{\hskip25pt usw}} & \multirow{-8}{*}{\raggedleft\arraybackslash 10} & \multirow{-3}{*}{\raggedleft\arraybackslash 100} & 0.9 & \cellcolor{gray!20}{\textbf{36.79}} & 0.00 & 4-5 & \cellcolor{gray!20}{\textbf{36.79}} & 0.00 & 4-5 & \cellcolor{gray!20}{\textbf{36.79}} & 0.00 & 4-5 & 36.53 & 0.08 & 5 & 34.95 & 0.00 & \\
\bottomrule
\end{tabular}\end{scriptsize}
\end{table*}

\section{Experimental Setup}
\label{sec:experimental_setup}

We now study the proposed evolutionary approaches on a set of benchmark instances. We first describe the experimental setup and discuss results afterwards.

Due to the lack of real-world knapsack instances we consider four types of benchmark instances with each $n=100$ items frequently used for benchmarking~\cite{PISINGER20052271,KelPfePis04}.
For \emph{strongly correlated} instances (scorr) the weight is sampled uniformly at random from $\{1, \ldots, R\}$ and for the profits $p_i = w_i + R/10$ holds where $R = 10\,000$. \emph{Inversely strongly correlated} instances (invscorr) are similar. Here, profits are sampled from $\{1, \ldots, R\}$ and weights correspond to profits plus $R/10$. For \emph{uncorrelated} instances both weights and profits are sampled from $\{1, \ldots, R\}$ at random, i.e., there is no correlation at all. Eventually, for instances of type \emph{uniform similar weights}~(usw) the weights are distributed in $\{100\,000, \ldots, 100\,100\}$ and profits are located in $\{1, \ldots, 1\,000\}$.
For each instance we consider $D \in \{2, 5, 10\}$ and set the knapsack capacity to
$W = \frac{D}{11} \cdot \sum_{i=1}^{n} w_i$.
Intuitively, the closer $W$ is to half the weight of all items, the more flexibility one would expect with respect to item packing while very low and very high values are rather inflexible. In particular for $D=10$ most items fit into the knapsack and the space of feasible solution may be small.
We study $\mu \in \{25, 50, 100, 200\}$ to cover the cases where the population size is less, equal to and larger than the instance size. In addition, we consider $\varepsilon \in \{0.1, 0.2, 0.5, 0.9\}$ to define the quality threshold. This parameter, in interaction with $D$ (i.e., implicitly $W$), should be crucial for the maximally possible population diversity as the combination $(\varepsilon, D)$ might strongly limit the number of feasible packings.
Moreover, we study the five different mutation operators (BF, PBF, HTBF, EDO-BBF1, EDO-BBF2), crossover (off, on with probability $p_c=0.8$) and repair of infeasible solutions (on/off). We run the $(\mu+1)$-EA for each combination of the cross-product of all parameters ($4\,608$ in total) ten times independently. In each experiment the population is initialized with $\mu$ copies of an $(1-\varepsilon/2)$-approximate solution calculated by the FPTAS. The EA is terminated after $\mu n$ iterations.
In the following we often use the term \emph{setup} to refer to a tuple $(\text{instance type}, D, \mu, \varepsilon)$ describing a combination of experimental parameters.

Our implementation is based on Python3.
Code, evaluation scripts and data are available in a public GitHub repository for the sake of reproducibility.\footnote{Code and data: \url{https://github.com/jakobbossek/GECCO2021-knapsack-diversity}.}

\section{Experimental Results}
\label{sec:experimental_results}

The section first considers the setting with a \emph{generous} budget of $\mu n$ function evaluations and discusses result for a \emph{restricted} budget of only $\mu$ iterations afterwards. Eventually the effect of crossover is investigated.

\subsection{Benchmark of mutation operators with generous budget}

We first compare the entropy values of the evolved populations of $(\mu+1)$-EA with mutation only and activated repair. Results without repair show the exact same trends; we will come back to repair later.
Table~\ref{tab:entropy_mutation_only_with_repair} shows the mean entropy and standard deviation for all five mutation operators separated by all considered parameters. We omitted results for $\mu\in\{50,200\}$, $\varepsilon \in \{0.2\}$ and instance type invscorr (almost identical to scorr) due to space limitations. The omitted data though does not reveal any further insights. We first observe that in general, with increasing $\varepsilon$, the achieved entropy values tend to increase given a fixed value of $D$. This was expected since with $\varepsilon \to 1$ the number of feasible solutions, i.e., such solutions that do not violate the minimal quality criterion, grows monotonically. However, this effect is far less pronounced for (inversely) strongly correlated instances which can be attributed to the strong correlation. Another interesting observation concerns the standard deviation. In fact, for PBF and HTBF in 90\% of the cases the standard deviation is below $0.006$ with $0.132$ and $0.125$ being the maximum values. Tailored operators are less robust, but even here the standard deviation is quite low (at most $0.271$ for EDO-BBF1, and $1.43$ for EDO-BBF2). The last general observation is due to runs that can be categorized as \enquote{failed}. Here, either all algorithms fail which occurs for $D=10$ and $\varepsilon \leq 0.2$ (here both parameters in combination hinder the algorithms from generating different feasible solutions) or (a part of) the biased mutation fails due to its working principles being inappropriate for the special setting. Note that EDO-BBF1 always fails for population size larger than $n$ as it is designed for $\mu < n$.

Besides the general observations we see a clear picture. After $\mu n$ iterations standard mutation operators (BF, PBF and HTBF) are superior with respect to mean entropy. These results are statistically supported by the results of pairwise Wilcoxon-Mann-Whitney tests at significance level $\alpha=0.05$ (see stat columns in the table) with Bonferroni-Holm correction to account for multiple-testing of samples. However, note that these results can be explained by the very low standard deviation. In fact, comparing the plain entropy numbers, all operators perform reasonably well in the majority of cases. Deviations from the maximum achieved mean entropy per setup are within $5.81\%$ in 99\% of the cases leaving apart some pathological cases where biased mutation does not work at all (recall that EDO-BBF1 does not work if $\mu > n$).
All in all, flipping multiple bits (with increased probability) in an unbiased way seems beneficial \underline{in the long term} as PBF mutation mostly performs best.

\begin{figure}
    \centering
    \includegraphics[width=\columnwidth]{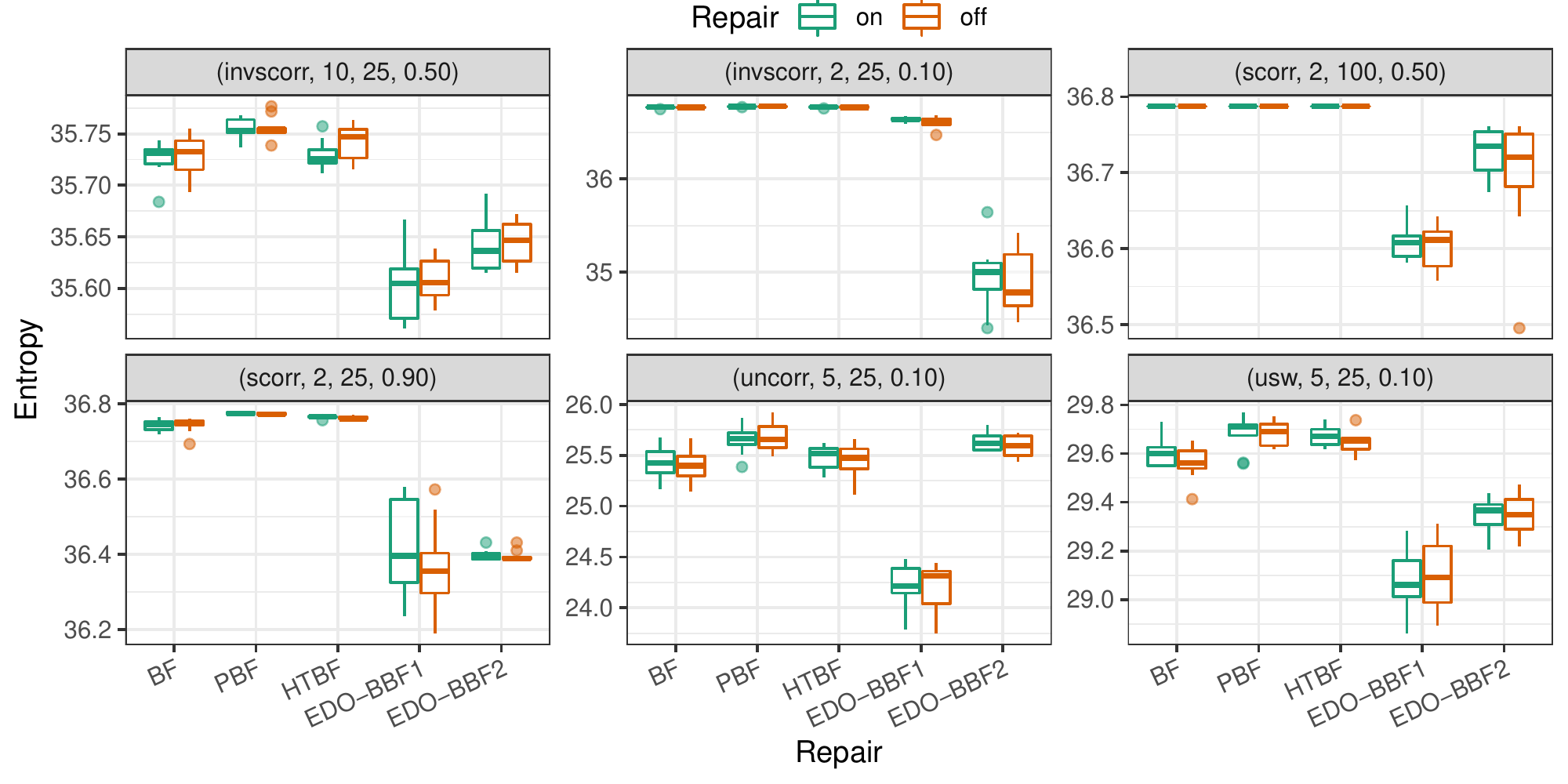}
    \vskip-10pt
    \caption{Distribution of population diversity (entropy) for six representative setups with and without repair operator.}
    \label{fig:entropy_boxplots_repair}
\end{figure}

The effect of active versus inactive repair is mostly minuscule. The overall observations drawn from the data with active repair directly carry over to the ones without. We therefore refrain from showing a page-filling table and instead visualize distributions for representative setups in Figure~\ref{fig:entropy_boxplots_repair}. This can be attributed to slightly more individuals being feasible after mutation.

\subsection{Benchmark of mutation operators with restricted budget}

\begin{figure}[ht]
    \centering
    \includegraphics[width=\linewidth]{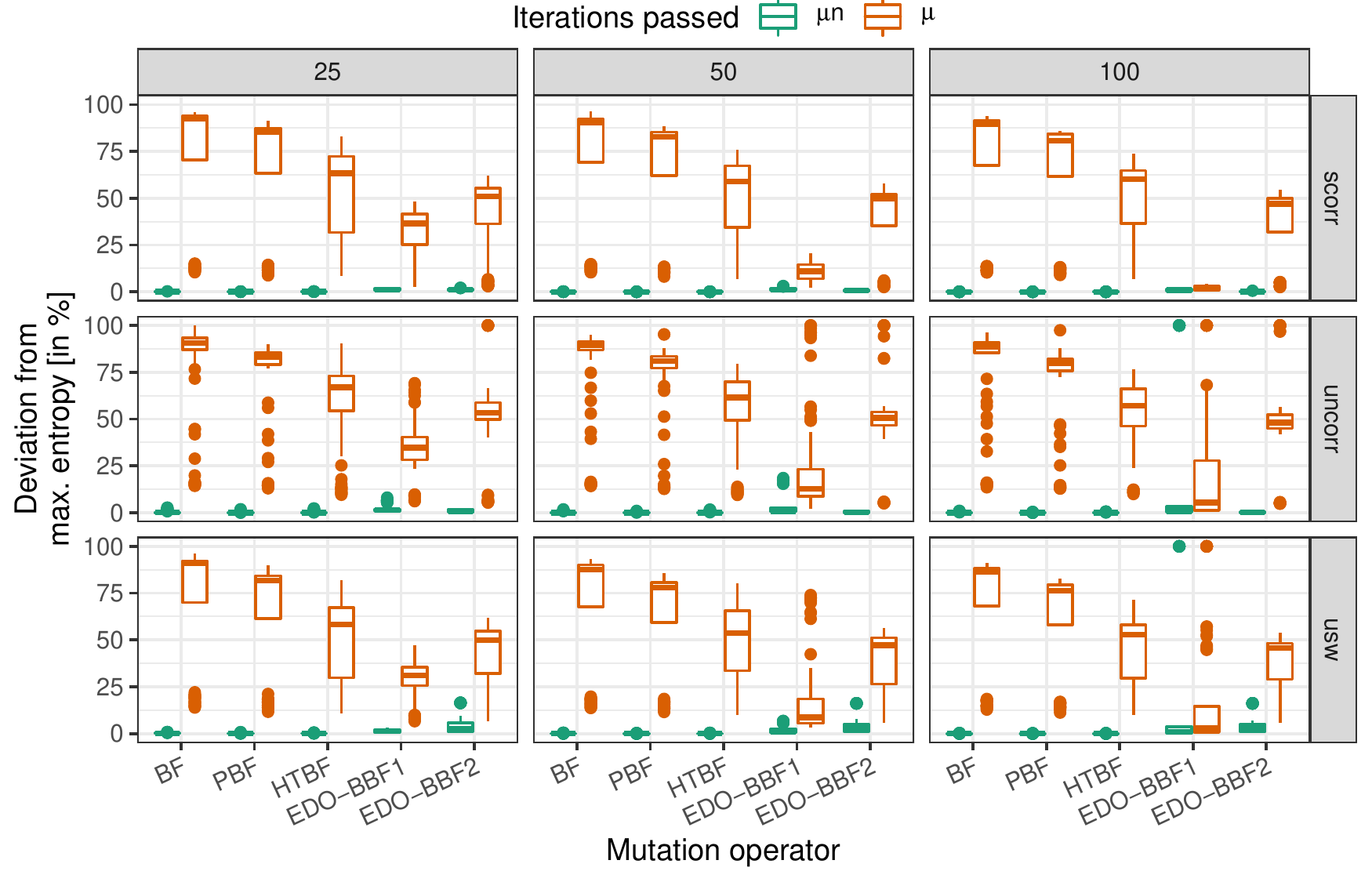}
    \vskip-10pt
    \caption{Distribution of the normalized distance to the maximum entropy achieved over all runs on a different instances after $\mu n$ iterations (actual budget for EA) and a severely restricted budget of only $\mu$ iterations. Data is split by instance type (rows) and $\mu$ (columns).}
    \label{fig:anytime_distance_to_max_entropy}
\end{figure}
\begin{figure}[ht]
    \centering
    \includegraphics[width=\columnwidth]{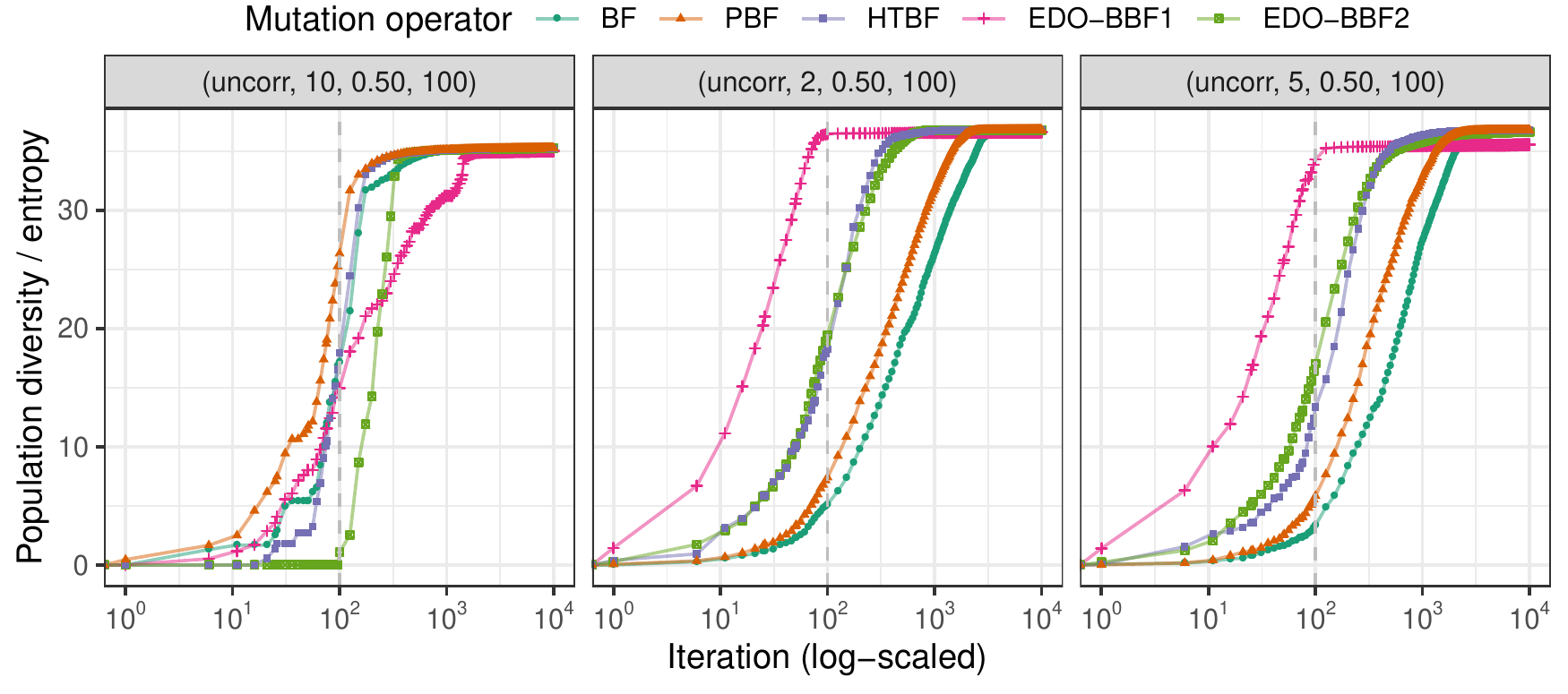}
    \vskip-10pt
    \caption{Trajectories of the population diversity in the course of iterations with focus on severely reduced budget. The plot headers read as follows: $(\text{type}, D, \varepsilon, \mu$).}
    \label{fig:traces}
\end{figure}
\begin{table*}[htbp]

\caption{\label{tab:}Mean (\text{mean}) and standard deviation (\text{std}) of mean entropy. Highest mean values are highlighted in \colorbox{gray!20}{\textbf{bold face}}. Algorithms are numbered (see second row).}
\label{tab:entropy_mutation_only_with_repair_reduced_budget}
\renewcommand{\arraystretch}{0.6}
\renewcommand{\tabcolsep}{7pt}
\centering
\begin{scriptsize}\begin{tabular}[t]{crrrrrrrrrrrrrrrrrr}
\toprule
\multicolumn{1}{c}{\textbf{ }} & \multicolumn{1}{c}{\textbf{ }} & \multicolumn{1}{c}{\textbf{ }} & \multicolumn{1}{c}{\textbf{ }} & \multicolumn{9}{c}{\textbf{Standard}} & \multicolumn{6}{c}{\textbf{Biased}} \\
\cmidrule(l{3pt}r{3pt}){5-13} \cmidrule(l{3pt}r{3pt}){14-19}
\multicolumn{1}{c}{\textbf{ }} & \multicolumn{1}{c}{\textbf{ }} & \multicolumn{1}{c}{\textbf{ }} & \multicolumn{1}{c}{\textbf{ }} & \multicolumn{3}{c}{\textbf{BF (1)}} & \multicolumn{3}{c}{\textbf{PBF (2)}} & \multicolumn{3}{c}{\textbf{HTBF (3)}} & \multicolumn{3}{c}{\textbf{EDO-BBF1 (4)}} & \multicolumn{3}{c}{\textbf{EDO-BBF2 (5)}} \\
\cmidrule(l{3pt}r{3pt}){5-7} \cmidrule(l{3pt}r{3pt}){8-10} \cmidrule(l{3pt}r{3pt}){11-13} \cmidrule(l{3pt}r{3pt}){14-16} \cmidrule(l{3pt}r{3pt}){17-19}
 & $D$ & $\mu$ & $\varepsilon$ & \textbf{mean} & \textbf{std} & \textbf{stat} & \textbf{mean} & \textbf{std} & \textbf{stat} & \textbf{mean} & \textbf{std} & \textbf{stat} & \textbf{mean} & \textbf{std} & \textbf{stat} & \textbf{mean} & \textbf{std} & \textbf{stat}\\
\midrule
 &  &  & 0.1 & 2.98 & 0.59 &  & 5.60 & 0.86 & 1 & 13.69 & 4.56 & 1-2 & \cellcolor{gray!20}{\textbf{23.24}} & 0.82 & 1-3,5 & 16.61 & 1.42 & 1-3\\

 &  &  & 0.5 & 2.84 & 0.49 &  & 5.64 & 0.57 & 1 & 12.40 & 2.84 & 1-2 & \cellcolor{gray!20}{\textbf{23.43}} & 1.29 & 1-3,5 & 16.82 & 1.79 & 1-3\\

 &  & \multirow{-3}{*}{\raggedleft\arraybackslash 25} & 0.9 & 2.93 & 0.75 &  & 5.32 & 0.23 & 1 & 13.10 & 4.03 & 1-2 & \cellcolor{gray!20}{\textbf{23.36}} & 0.89 & 1-3,5 & 16.64 & 1.73 & 1-3\\

\cmidrule{3-19}
 &  &  & 0.1 & 4.06 & 0.71 &  & 7.34 & 0.45 & 1 & 15.12 & 1.80 & 1-2 & \cellcolor{gray!20}{\textbf{36.37}} & 0.07 & 1-3,5 & 19.28 & 1.23 & 1-3\\

 &  &  & 0.5 & 3.86 & 0.62 &  & 7.43 & 0.26 & 1 & 15.28 & 2.58 & 1-2 & \cellcolor{gray!20}{\textbf{36.45}} & 0.06 & 1-3,5 & 19.00 & 0.89 & 1-3\\

 & \multirow{-8}{*}{\raggedleft\arraybackslash 2} & \multirow{-3}{*}{\raggedleft\arraybackslash 100} & 0.9 & 4.17 & 0.60 &  & 7.27 & 0.51 & 1 & 14.88 & 2.37 & 1-2 & \cellcolor{gray!20}{\textbf{36.42}} & 0.06 & 1-3,5 & 18.66 & 1.00 & 1-3\\

\cmidrule{2-19}
 &  &  & 0.1 & 2.33 & 0.58 &  & 4.85 & 0.76 & 1 & 10.75 & 2.19 & 1-2 & \cellcolor{gray!20}{\textbf{21.83}} & 1.96 & 1-3,5 & 17.49 & 1.24 & 1-3\\

 &  &  & 0.5 & 2.27 & 0.57 &  & 4.49 & 0.70 & 1 & 12.36 & 3.50 & 1-2 & \cellcolor{gray!20}{\textbf{21.03}} & 1.22 & 1-3,5 & 17.19 & 1.73 & 1-3\\

 &  & \multirow{-3}{*}{\raggedleft\arraybackslash 25} & 0.9 & 2.15 & 0.44 &  & 4.06 & 0.42 & 1 & 10.04 & 3.21 & 1-2 & \cellcolor{gray!20}{\textbf{20.82}} & 0.95 & 1-3,5 & 17.70 & 1.13 & 1-3\\

\cmidrule{3-19}
 &  &  & 0.1 & 3.05 & 0.37 &  & 5.59 & 0.37 & 1 & 13.00 & 2.23 & 1-2 & \cellcolor{gray!20}{\textbf{35.41}} & 0.06 & 1-3,5 & 18.72 & 1.19 & 1-3\\

 &  &  & 0.5 & 3.07 & 0.46 &  & 5.59 & 0.41 & 1 & 12.94 & 2.00 & 1-2 & \cellcolor{gray!20}{\textbf{35.46}} & 0.08 & 1-3,5 & 19.33 & 1.47 & 1-3\\

 & \multirow{-8}{*}{\raggedleft\arraybackslash 5} & \multirow{-3}{*}{\raggedleft\arraybackslash 100} & 0.9 & 2.95 & 0.43 &  & 5.58 & 0.29 & 1 & 12.36 & 1.70 & 1-2 & \cellcolor{gray!20}{\textbf{35.48}} & 0.07 & 1-3,5 & 18.58 & 1.16 & 1-3\\

\cmidrule{2-19}
 &  &  & 0.1 & \cellcolor{gray!20}{\textbf{0.00}} & 0.00 &  & \cellcolor{gray!20}{\textbf{0.00}} & 0.00 &  & \cellcolor{gray!20}{\textbf{0.00}} & 0.00 &  & \cellcolor{gray!20}{\textbf{0.00}} & 0.00 &  & \cellcolor{gray!20}{\textbf{0.00}} & 0.00 & \\

 &  &  & 0.5 & 31.65 & 0.28 &  & 31.97 & 0.44 &  & 32.54 & 0.25 & 1-2 & 34.34 & 0.47 & 1-3 & \cellcolor{gray!20}{\textbf{34.65}} & 0.12 & 1-3\\

 &  & \multirow{-3}{*}{\raggedleft\arraybackslash 25} & 0.9 & 31.66 & 0.33 &  & 31.84 & 0.32 &  & 32.28 & 0.54 & 1-2 & 33.77 & 0.41 & 1-3 & \cellcolor{gray!20}{\textbf{34.57}} & 0.28 & 1-4\\

\cmidrule{3-19}
 &  &  & 0.1 & \cellcolor{gray!20}{\textbf{0.00}} & 0.00 &  & \cellcolor{gray!20}{\textbf{0.00}} & 0.00 &  & \cellcolor{gray!20}{\textbf{0.00}} & 0.00 &  & \cellcolor{gray!20}{\textbf{0.00}} & 0.00 &  & \cellcolor{gray!20}{\textbf{0.00}} & 0.00 & \\

 &  &  & 0.5 & 31.92 & 0.17 &  & 32.35 & 0.23 & 1 & 33.10 & 0.26 & 1-2 & \cellcolor{gray!20}{\textbf{35.15}} & 0.11 & 1-3,5 & 34.92 & 0.06 & 1-3\\

\multirow{-20}{*}{\centering\arraybackslash \rotatebox[origin=c]{90}{\hskip25pt scorr}} & \multirow{-8}{*}{\raggedleft\arraybackslash 10} & \multirow{-3}{*}{\raggedleft\arraybackslash 100} & 0.9 & 31.94 & 0.24 &  & 32.21 & 0.21 & 1 & 32.88 & 0.36 & 1-2 & \cellcolor{gray!20}{\textbf{36.25}} & 0.08 & 1-3,5 & 34.96 & 0.04 & 1-3\\
\cmidrule{1-19}
\cmidrule{3-19}
 &  &  & 0.1 & 3.08 & 0.98 &  & 6.27 & 0.90 & 1 & 14.06 & 5.11 & 1-2 & \cellcolor{gray!20}{\textbf{24.55}} & 1.52 & 1-3,5 & 17.44 & 2.12 & 1-3\\

 &  &  & 0.5 & 3.30 & 0.33 &  & 6.08 & 0.50 & 1 & 12.49 & 2.50 & 1-2 & \cellcolor{gray!20}{\textbf{24.87}} & 1.48 & 1-3,5 & 17.07 & 1.57 & 1-3\\

 &  & \multirow{-3}{*}{\raggedleft\arraybackslash 25} & 0.9 & 3.52 & 1.06 &  & 6.32 & 1.08 & 1 & 11.95 & 2.08 & 1-2 & \cellcolor{gray!20}{\textbf{24.71}} & 0.60 & 1-3,5 & 16.98 & 1.05 & 1-3\\

\cmidrule{3-19}
 &  &  & 0.1 & 4.27 & 0.53 &  & 7.26 & 0.38 & 1 & 16.16 & 1.48 & 1-2 & \cellcolor{gray!20}{\textbf{30.67}} & 0.96 & 1-3,5 & 19.95 & 0.83 & 1-3\\

 &  &  & 0.5 & 4.08 & 0.61 &  & 7.61 & 0.36 & 1 & 16.30 & 2.47 & 1-2 & \cellcolor{gray!20}{\textbf{36.44}} & 0.07 & 1-3,5 & 19.77 & 0.47 & 1-3\\

 & \multirow{-8}{*}{\raggedleft\arraybackslash 2} & \multirow{-3}{*}{\raggedleft\arraybackslash 100} & 0.9 & 4.30 & 0.48 &  & 7.44 & 0.49 & 1 & 15.08 & 2.50 & 1-2 & \cellcolor{gray!20}{\textbf{36.49}} & 0.06 & 1-3,5 & 19.59 & 1.06 & 1-3\\

\cmidrule{2-19}
 &  &  & 0.1 & 2.55 & 0.57 &  & 4.52 & 0.71 & 1 & 6.87 & 1.24 & 1-2 & 10.93 & 1.53 & 1-3 & \cellcolor{gray!20}{\textbf{12.47}} & 1.11 & 1-4\\

 &  &  & 0.5 & 2.31 & 0.65 &  & 4.98 & 0.83 & 1 & 10.22 & 3.33 & 1-2 & \cellcolor{gray!20}{\textbf{23.03}} & 1.20 & 1-3,5 & 16.81 & 1.50 & 1-3\\

 &  & \multirow{-3}{*}{\raggedleft\arraybackslash 25} & 0.9 & 2.29 & 0.54 &  & 4.78 & 0.44 & 1 & 11.24 & 3.96 & 1-2 & \cellcolor{gray!20}{\textbf{22.64}} & 1.96 & 1-3,5 & 16.06 & 1.53 & 1-3\\

\cmidrule{3-19}
 &  &  & 0.1 & 3.41 & 0.48 & 4 & 6.20 & 0.51 & 1,4 & 7.11 & 0.53 & 1-2,4 & 0.00 & 0.00 &  & \cellcolor{gray!20}{\textbf{13.46}} & 0.75 & 1-4\\

 &  &  & 0.5 & 3.29 & 0.20 &  & 5.97 & 0.46 & 1 & 14.26 & 3.02 & 1-2 & \cellcolor{gray!20}{\textbf{33.84}} & 0.37 & 1-3,5 & 17.67 & 0.62 & 1-3\\

 & \multirow{-8}{*}{\raggedleft\arraybackslash 5} & \multirow{-3}{*}{\raggedleft\arraybackslash 100} & 0.9 & 3.19 & 0.41 &  & 6.01 & 0.50 & 1 & 12.61 & 1.47 & 1-2 & \cellcolor{gray!20}{\textbf{35.35}} & 0.08 & 1-3,5 & 17.48 & 0.87 & 1-3\\

\cmidrule{2-19}
 &  &  & 0.1 & \cellcolor{gray!20}{\textbf{0.00}} & 0.00 &  & \cellcolor{gray!20}{\textbf{0.00}} & 0.00 &  & \cellcolor{gray!20}{\textbf{0.00}} & 0.00 &  & \cellcolor{gray!20}{\textbf{0.00}} & 0.00 &  & \cellcolor{gray!20}{\textbf{0.00}} & 0.00 & \\

 &  &  & 0.5 & 14.53 & 9.71 & 5 & 16.95 & 9.09 & 5 & 18.97 & 9.73 & 5 & \cellcolor{gray!20}{\textbf{19.79}} & 6.45 & 5 & 1.18 & 3.72 & \\

 &  & \multirow{-3}{*}{\raggedleft\arraybackslash 25} & 0.9 & 30.91 & 0.54 &  & 31.41 & 0.28 & 1 & 32.28 & 0.64 & 1-2 & 33.70 & 0.49 & 1-3 & \cellcolor{gray!20}{\textbf{34.32}} & 0.53 & 1-4\\

\cmidrule{3-19}
 &  &  & 0.1 & \cellcolor{gray!20}{\textbf{0.00}} & 0.00 &  & \cellcolor{gray!20}{\textbf{0.00}} & 0.00 &  & \cellcolor{gray!20}{\textbf{0.00}} & 0.00 &  & \cellcolor{gray!20}{\textbf{0.00}} & 0.00 &  & \cellcolor{gray!20}{\textbf{0.00}} & 0.00 & \\

 &  &  & 0.5 & 13.64 & 7.46 & 5 & 14.95 & 9.01 & 5 & \cellcolor{gray!20}{\textbf{22.50}} & 3.23 & 1-2,4-5 & 13.43 & 1.69 & 5 & 0.22 & 0.47 & \\

\multirow{-20}{*}{\centering\arraybackslash \rotatebox[origin=c]{90}{\hskip25pt uncorr}} & \multirow{-8}{*}{\raggedleft\arraybackslash 10} & \multirow{-3}{*}{\raggedleft\arraybackslash 100} & 0.9 & 31.24 & 0.26 &  & 31.75 & 0.23 & 1 & 32.65 & 0.28 & 1-2 & \cellcolor{gray!20}{\textbf{36.21}} & 0.10 & 1-3,5 & 34.90 & 0.09 & 1-3\\
\cmidrule{1-19}
\cmidrule{3-19}
 &  &  & 0.1 & 3.43 & 0.60 &  & 6.51 & 1.05 & 1 & 16.62 & 4.57 & 1-2 & \cellcolor{gray!20}{\textbf{26.55}} & 1.36 & 1-3,5 & 21.41 & 2.05 & 1-3\\

 &  &  & 0.5 & 3.42 & 0.78 &  & 6.28 & 0.64 & 1 & 14.20 & 3.47 & 1-2 & \cellcolor{gray!20}{\textbf{26.06}} & 0.94 & 1-3,5 & 16.78 & 1.18 & 1-2\\

 &  & \multirow{-3}{*}{\raggedleft\arraybackslash 25} & 0.9 & 3.69 & 0.68 &  & 6.78 & 0.65 & 1 & 12.69 & 3.33 & 1-2 & \cellcolor{gray!20}{\textbf{25.63}} & 1.03 & 1-3,5 & 16.94 & 1.66 & 1-3\\

\cmidrule{3-19}
 &  &  & 0.1 & 4.89 & 0.32 &  & 8.62 & 0.59 & 1 & 17.25 & 1.89 & 1-2 & \cellcolor{gray!20}{\textbf{36.54}} & 0.02 & 1-3,5 & 24.25 & 0.76 & 1-3\\

 &  &  & 0.5 & 5.30 & 0.41 &  & 8.44 & 0.55 & 1 & 18.26 & 2.00 & 1-2 & \cellcolor{gray!20}{\textbf{36.55}} & 0.04 & 1-3,5 & 19.51 & 0.79 & 1-2\\

 & \multirow{-8}{*}{\raggedleft\arraybackslash 2} & \multirow{-3}{*}{\raggedleft\arraybackslash 100} & 0.9 & 5.02 & 0.31 &  & 9.03 & 0.87 & 1 & 18.48 & 2.23 & 1-2 & \cellcolor{gray!20}{\textbf{36.57}} & 0.04 & 1-3,5 & 19.06 & 0.63 & 1-2\\

\cmidrule{2-19}
 &  &  & 0.1 & 2.48 & 0.71 &  & 5.80 & 0.89 & 1 & 9.61 & 1.82 & 1-2 & \cellcolor{gray!20}{\textbf{17.26}} & 0.78 & 1-3,5 & 16.11 & 1.55 & 1-3\\

 &  &  & 0.5 & 2.72 & 0.72 &  & 5.67 & 0.61 & 1 & 13.09 & 2.68 & 1-2 & \cellcolor{gray!20}{\textbf{23.81}} & 1.30 & 1-3,5 & 16.79 & 1.23 & 1-3\\

 &  & \multirow{-3}{*}{\raggedleft\arraybackslash 25} & 0.9 & 2.76 & 0.46 &  & 5.48 & 0.68 & 1 & 13.82 & 3.70 & 1-2 & \cellcolor{gray!20}{\textbf{24.05}} & 1.60 & 1-3,5 & 15.98 & 1.35 & 1-3\\

\cmidrule{3-19}
 &  &  & 0.1 & 3.79 & 0.45 & 4 & 7.15 & 0.54 & 1,4 & 10.13 & 1.18 & 1-2,4 & 0.00 & 0.00 &  & \cellcolor{gray!20}{\textbf{16.50}} & 0.55 & 1-4\\

 &  &  & 0.5 & 4.06 & 0.46 &  & 6.96 & 0.36 & 1 & 14.69 & 2.50 & 1-2 & \cellcolor{gray!20}{\textbf{35.24}} & 0.05 & 1-3,5 & 18.36 & 1.06 & 1-3\\

 & \multirow{-8}{*}{\raggedleft\arraybackslash 5} & \multirow{-3}{*}{\raggedleft\arraybackslash 100} & 0.9 & 3.85 & 0.35 &  & 7.04 & 0.40 & 1 & 15.30 & 2.20 & 1-2 & \cellcolor{gray!20}{\textbf{35.28}} & 0.05 & 1-3,5 & 18.59 & 0.79 & 1-3\\

\cmidrule{2-19}
 &  &  & 0.1 & \cellcolor{gray!20}{\textbf{0.00}} & 0.00 &  & \cellcolor{gray!20}{\textbf{0.00}} & 0.00 &  & \cellcolor{gray!20}{\textbf{0.00}} & 0.00 &  & \cellcolor{gray!20}{\textbf{0.00}} & 0.00 &  & \cellcolor{gray!20}{\textbf{0.00}} & 0.00 & \\

 &  &  & 0.5 & 29.63 & 0.67 & 4-5 & 30.17 & 0.45 & 1,4-5 & \cellcolor{gray!20}{\textbf{30.37}} & 1.08 & 4-5 & 26.88 & 1.77 &  & 28.64 & 1.39 & 4\\

 &  & \multirow{-3}{*}{\raggedleft\arraybackslash 25} & 0.9 & 29.57 & 0.51 &  & 30.20 & 0.51 & 1 & 31.33 & 0.62 & 1-2 & 33.48 & 0.51 & 1-3 & \cellcolor{gray!20}{\textbf{34.06}} & 0.33 & 1-4\\

\cmidrule{3-19}
 &  &  & 0.1 & \cellcolor{gray!20}{\textbf{0.00}} & 0.00 &  & \cellcolor{gray!20}{\textbf{0.00}} & 0.00 &  & \cellcolor{gray!20}{\textbf{0.00}} & 0.00 &  & \cellcolor{gray!20}{\textbf{0.00}} & 0.00 &  & \cellcolor{gray!20}{\textbf{0.00}} & 0.00 & \\

 &  &  & 0.5 & 30.24 & 0.27 & 4-5 & 30.53 & 0.35 & 1,4-5 & \cellcolor{gray!20}{\textbf{31.00}} & 0.54 & 1-2,4-5 & 17.55 & 1.79 &  & 28.13 & 1.58 & 4\\

\multirow{-20}{*}{\centering\arraybackslash \rotatebox[origin=c]{90}{\hskip25pt usw}} & \multirow{-8}{*}{\raggedleft\arraybackslash 10} & \multirow{-3}{*}{\raggedleft\arraybackslash 100} & 0.9 & 30.29 & 0.17 &  & 30.83 & 0.21 & 1 & 31.91 & 0.29 & 1-2 & \cellcolor{gray!20}{\textbf{36.11}} & 0.10 & 1-3,5 & 34.59 & 0.10 & 1-3\\
\bottomrule
\end{tabular}\end{scriptsize}
\end{table*}
We now turn our focus towards a severely restricted budget of function evaluations of just $\mu$. Note that the total budget considered in the previous investigation was based on a budget of $\mu n$ evaluations which is by a factor of $n$ (here $n=100$) higher. As a performance indicator we consider the percentage deviation (lower is better) from the maximum achieved population diversity over all runs of all algorithms on each setup $(\text{type}, \mu, \varepsilon, D)$. Figure~\ref{fig:anytime_distance_to_max_entropy} shows boxplots of the percent deviation from the maximum diversity split by instance type (scorr, uncorr and usw) and population size $\mu \in \{25, 50, 100\}$. In order to save precious space, we do not further split by $\varepsilon$ and $D$. However, the trends observed from Figure~\ref{fig:anytime_distance_to_max_entropy} are stable across all considered setups. We observe a very clear trend/ranking respectively. Considering the unbiased operators -- with respect to median deviation, BF is always outperformed by PBF which in turn is always outperformed by HTBF in the setting with strongly reduced budget. Switching to he biased operators both are always superior to HTBF with EDO-BBF1 outperforming EDO-BBF2 consistently and statistically significant. EDO-BBF1 achieves median deviations of less then 32\% for $\mu=25$, less then 12.5\% for $\mu=50$ and even values below 10\% for $\mu=100$. This means there is clear trend towards better \emph{short-term} performance with $\mu$ approaching $n$. Biased mutation also shows lower variance. In particular for (inversely) strongly correlated instances the boxplots of EDO-BBF1 get very narrow.
In total, EDO-BBF1 show the best median performance in $80.56\%$ of the cases, EDO-BBF2 is placed second with $12.50\%$ and HTBF in $6.94\%$; BF and PBF score first in none of the cases. Table~\ref{tab:entropy_mutation_only_with_repair_reduced_budget} gives a detailed and less aggregated overview in terms of numbers and results of significance tests following the style of Table~\ref{tab:entropy_mutation_only_with_repair}. The observation derived from the boxplots are confirmed.
Figure~\ref{fig:traces} gives a visual impression of actual runs of the $(\mu+1)$-EA for different representative setups. Here, we clearly see the advantage of biased mutation after only $\mu$ iterations and the fast progress (note that the $x$-axis is on log-scale in order to make the first $\mu$ iterations more visible).

\subsection{Impact of crossover}

The results for crossover are mixed. Table~\ref{alg:crossover} shows a comparison of the crossover version~(CO) and mutation-only version~(no-CO) for the mutation operators BF, PBF and the biased EDO-BBF1. We observe that crossover often supports the biased mutation in particular for increasing $\varepsilon$, but often has the reverse effect for standard mutation. These observations hold true for both the generous and restricted budget scenarios. A reasonable explanation for this effect is that biased mutation likely is stuck in local optima close to the optimum where the strong bias prevents the algorithm from making further progress. Here, crossover might help to overcome these plateaus.

\begin{table*}[htbp]

\caption{\label{tab:}Mean (\text{mean}), standard deviation (\text{std}) and results of Wilcoxon-Mann-Whitney tests at a significance level of $\alpha = 0.05$ (\textbf{stat}) in terms of mean entropy for three mutation operators with active crossover (CO) and inactive crossover (noCO). Highest entropy values are highlighted in \colorbox{gray!20}{\textbf{bold face}}.}
\label{tab:crossover_comparison}
\renewcommand{\arraystretch}{0.58}
\renewcommand{\tabcolsep}{5.6pt}
\centering
\begin{scriptsize}\begin{tabular}[t]{crrr|rrrrrr|rrrrrr|rrrrrr}
\toprule
\multicolumn{1}{c}{\textbf{ }} & \multicolumn{1}{c}{\textbf{ }} & \multicolumn{1}{c}{\textbf{ }} & \multicolumn{1}{c}{\textbf{ }} & \multicolumn{6}{c}{\textbf{BF}} & \multicolumn{6}{c}{\textbf{PBF}} & \multicolumn{6}{c}{\textbf{EDO-BBF1}} \\
\cmidrule(l{3pt}r{3pt}){5-10} \cmidrule(l{3pt}r{3pt}){11-16} \cmidrule(l{3pt}r{3pt}){17-22}
\multicolumn{1}{c}{\textbf{ }} & \multicolumn{1}{c}{\textbf{ }} & \multicolumn{1}{c}{\textbf{ }} & \multicolumn{1}{c}{\textbf{ }} & \multicolumn{3}{c}{\textbf{no-CO (1)}} & \multicolumn{3}{c}{\textbf{CO (2)}} & \multicolumn{3}{c}{\textbf{no-CO (1)}} & \multicolumn{3}{c}{\textbf{CO (2)}} & \multicolumn{3}{c}{\textbf{no-CO (1)}} & \multicolumn{3}{c}{\textbf{CO (2)}} \\
\cmidrule(l{3pt}r{3pt}){5-7} \cmidrule(l{3pt}r{3pt}){8-10} \cmidrule(l{3pt}r{3pt}){11-13} \cmidrule(l{3pt}r{3pt}){14-16} \cmidrule(l{3pt}r{3pt}){17-19} \cmidrule(l{3pt}r{3pt}){20-22}
 & $D$ & $\mu$ & $\varepsilon$ & \textbf{mean} & \textbf{std} & \textbf{stat} & \textbf{mean} & \textbf{std} & \textbf{stat} & \textbf{mean} & \textbf{std} & \textbf{stat} & \textbf{mean} & \textbf{std} & \textbf{stat} & \textbf{mean} & \textbf{std} & \textbf{stat} & \textbf{mean} & \textbf{std} & \textbf{stat}\\
\midrule
 &  &  & 0.1 & \cellcolor{gray!20}{\textbf{36.74}} & 0.02 & \textcolor{black}{$\text{2}^{+}$} & 36.64 & 0.06 & $\text{1}^{-}$ & \cellcolor{gray!20}{\textbf{36.77}} & 0.00 & \textcolor{black}{$\text{2}^{+}$} & 36.74 & 0.01 & $\text{1}^{-}$ & 36.36 & 0.11 & $\text{2}^{-}$ & \cellcolor{gray!20}{\textbf{36.75}} & 0.01 & \textcolor{black}{$\text{1}^{+}$}\\

 &  &  & 0.5 & \cellcolor{gray!20}{\textbf{36.75}} & 0.01 & \textcolor{black}{$\text{2}^{+}$} & 36.22 & 0.22 & $\text{1}^{-}$ & \cellcolor{gray!20}{\textbf{36.77}} & 0.00 & \textcolor{black}{$\text{2}^{+}$} & 36.69 & 0.03 & $\text{1}^{-}$ & 36.38 & 0.10 & $\text{2}^{-}$ & \cellcolor{gray!20}{\textbf{36.76}} & 0.01 & \textcolor{black}{$\text{1}^{+}$}\\

 &  & \multirow{-3}{*}{\raggedleft\arraybackslash 25} & 0.9 & \cellcolor{gray!20}{\textbf{36.74}} & 0.02 & \textcolor{black}{$\text{2}^{+}$} & 35.38 & 0.45 & $\text{1}^{-}$ & \cellcolor{gray!20}{\textbf{36.77}} & 0.00 & \textcolor{black}{$\text{2}^{+}$} & 36.62 & 0.06 & $\text{1}^{-}$ & 36.42 & 0.13 & $\text{2}^{-}$ & \cellcolor{gray!20}{\textbf{36.75}} & 0.01 & \textcolor{black}{$\text{1}^{+}$}\\

\cmidrule{3-22}
 &  &  & 0.1 & \cellcolor{gray!20}{\textbf{36.79}} & 0.00 & \textcolor{black}{$\text{2}^{+}$} & \cellcolor{gray!20}{\textbf{36.79}} & 0.00 & $\text{1}^{-}$ & \cellcolor{gray!20}{\textbf{36.79}} & 0.00 & \textcolor{black}{$\text{2}^{+}$} & \cellcolor{gray!20}{\textbf{36.79}} & 0.00 & $\text{1}^{-}$ & 36.58 & 0.02 & $\text{2}^{-}$ & \cellcolor{gray!20}{\textbf{36.78}} & 0.00 & \textcolor{black}{$\text{1}^{+}$}\\

 &  &  & 0.5 & \cellcolor{gray!20}{\textbf{36.79}} & 0.00 & \textcolor{black}{$\text{2}^{+}$} & 36.78 & 0.00 & $\text{1}^{-}$ & \cellcolor{gray!20}{\textbf{36.79}} & 0.00 & \textcolor{black}{$\text{2}^{+}$} & \cellcolor{gray!20}{\textbf{36.79}} & 0.00 & $\text{1}^{-}$ & 36.61 & 0.02 & $\text{2}^{-}$ & \cellcolor{gray!20}{\textbf{36.78}} & 0.00 & \textcolor{black}{$\text{1}^{+}$}\\

 & \multirow{-10}{*}{\raggedleft\arraybackslash 2} & \multirow{-3}{*}{\raggedleft\arraybackslash 100} & 0.9 & \cellcolor{gray!20}{\textbf{36.79}} & 0.00 & \textcolor{black}{$\text{2}^{+}$} & 36.73 & 0.03 & $\text{1}^{-}$ & \cellcolor{gray!20}{\textbf{36.79}} & 0.00 & \textcolor{black}{$\text{2}^{+}$} & \cellcolor{gray!20}{\textbf{36.79}} & 0.00 & $\text{1}^{-}$ & 36.60 & 0.02 & $\text{2}^{-}$ & \cellcolor{gray!20}{\textbf{36.78}} & 0.00 & \textcolor{black}{$\text{1}^{+}$}\\

\cmidrule{2-22}
 &  &  & 0.1 & \cellcolor{gray!20}{\textbf{36.06}} & 0.02 & \textcolor{black}{$\text{2}^{+}$} & 36.00 & 0.01 & $\text{1}^{-}$ & \cellcolor{gray!20}{\textbf{36.10}} & 0.01 & \textcolor{black}{$\text{2}^{+}$} & 36.04 & 0.02 & $\text{1}^{-}$ & \cellcolor{gray!20}{\textbf{35.85}} & 0.03 & $\text{2}^{-}$ & 35.84 & 0.05 & $\text{1}^{-}$\\

 &  &  & 0.5 & 36.75 & 0.01 & $\text{2}^{-}$ & \cellcolor{gray!20}{\textbf{36.77}} & 0.00 & \textcolor{black}{$\text{1}^{+}$} & \cellcolor{gray!20}{\textbf{36.77}} & 0.00 & $\text{2}^{-}$ & \cellcolor{gray!20}{\textbf{36.77}} & 0.00 & $\text{1}^{-}$ & 36.11 & 0.08 & $\text{2}^{-}$ & \cellcolor{gray!20}{\textbf{36.76}} & 0.01 & \textcolor{black}{$\text{1}^{+}$}\\

 &  & \multirow{-3}{*}{\raggedleft\arraybackslash 25} & 0.9 & \cellcolor{gray!20}{\textbf{36.75}} & 0.01 & \textcolor{black}{$\text{2}^{+}$} & 36.51 & 0.09 & $\text{1}^{-}$ & \cellcolor{gray!20}{\textbf{36.77}} & 0.00 & \textcolor{black}{$\text{2}^{+}$} & 36.74 & 0.02 & $\text{1}^{-}$ & 36.08 & 0.08 & $\text{2}^{-}$ & \cellcolor{gray!20}{\textbf{36.76}} & 0.01 & \textcolor{black}{$\text{1}^{+}$}\\

\cmidrule{3-22}
 &  &  & 0.1 & \cellcolor{gray!20}{\textbf{36.17}} & 0.00 & \textcolor{black}{$\text{2}^{+}$} & 36.12 & 0.01 & $\text{1}^{-}$ & \cellcolor{gray!20}{\textbf{36.17}} & 0.00 & \textcolor{black}{$\text{2}^{+}$} & 36.14 & 0.01 & $\text{1}^{-}$ & 35.80 & 0.02 & $\text{2}^{-}$ & \cellcolor{gray!20}{\textbf{35.83}} & 0.02 & \textcolor{black}{$\text{1}^{+}$}\\

 &  &  & 0.5 & \cellcolor{gray!20}{\textbf{36.79}} & 0.00 & \textcolor{black}{$\text{2}^{+}$} & \cellcolor{gray!20}{\textbf{36.79}} & 0.00 & $\text{1}^{-}$ & \cellcolor{gray!20}{\textbf{36.79}} & 0.00 & \textcolor{black}{$\text{2}^{+}$} & \cellcolor{gray!20}{\textbf{36.79}} & 0.00 & $\text{1}^{-}$ & 35.87 & 0.02 & $\text{2}^{-}$ & \cellcolor{gray!20}{\textbf{36.77}} & 0.00 & \textcolor{black}{$\text{1}^{+}$}\\

 & \multirow{-10}{*}{\raggedleft\arraybackslash 5} & \multirow{-3}{*}{\raggedleft\arraybackslash 100} & 0.9 & \cellcolor{gray!20}{\textbf{36.79}} & 0.00 & \textcolor{black}{$\text{2}^{+}$} & \cellcolor{gray!20}{\textbf{36.79}} & 0.00 & $\text{1}^{-}$ & \cellcolor{gray!20}{\textbf{36.79}} & 0.00 & \textcolor{black}{$\text{2}^{+}$} & \cellcolor{gray!20}{\textbf{36.79}} & 0.00 & $\text{1}^{-}$ & 35.86 & 0.04 & $\text{2}^{-}$ & \cellcolor{gray!20}{\textbf{36.78}} & 0.00 & \textcolor{black}{$\text{1}^{+}$}\\

\cmidrule{2-22}
 &  &  & 0.1 & \cellcolor{gray!20}{\textbf{0.00}} & 0.00 & $\text{2}^{-}$ & \cellcolor{gray!20}{\textbf{0.00}} & 0.00 & $\text{1}^{-}$ & \cellcolor{gray!20}{\textbf{0.00}} & 0.00 & $\text{2}^{-}$ & \cellcolor{gray!20}{\textbf{0.00}} & 0.00 & $\text{1}^{-}$ & \cellcolor{gray!20}{\textbf{0.00}} & 0.00 & $\text{2}^{-}$ & \cellcolor{gray!20}{\textbf{0.00}} & 0.00 & $\text{1}^{-}$\\

 &  &  & 0.5 & \cellcolor{gray!20}{\textbf{35.85}} & 0.02 & \textcolor{black}{$\text{2}^{+}$} & 35.80 & 0.02 & $\text{1}^{-}$ & \cellcolor{gray!20}{\textbf{35.88}} & 0.01 & \textcolor{black}{$\text{2}^{+}$} & 35.80 & 0.02 & $\text{1}^{-}$ & \cellcolor{gray!20}{\textbf{35.68}} & 0.05 & \textcolor{black}{$\text{2}^{+}$} & 35.62 & 0.03 & $\text{1}^{-}$\\

 &  & \multirow{-3}{*}{\raggedleft\arraybackslash 25} & 0.9 & \cellcolor{gray!20}{\textbf{36.77}} & 0.01 & $\text{2}^{-}$ & \cellcolor{gray!20}{\textbf{36.77}} & 0.00 & \textcolor{black}{$\text{1}^{+}$} & \cellcolor{gray!20}{\textbf{36.77}} & 0.00 & $\text{2}^{-}$ & \cellcolor{gray!20}{\textbf{36.77}} & 0.00 & $\text{1}^{-}$ & 36.17 & 0.10 & $\text{2}^{-}$ & \cellcolor{gray!20}{\textbf{36.76}} & 0.01 & \textcolor{black}{$\text{1}^{+}$}\\

\cmidrule{3-22}
 &  &  & 0.1 & \cellcolor{gray!20}{\textbf{0.00}} & 0.00 & $\text{2}^{-}$ & \cellcolor{gray!20}{\textbf{0.00}} & 0.00 & $\text{1}^{-}$ & \cellcolor{gray!20}{\textbf{0.00}} & 0.00 & $\text{2}^{-}$ & \cellcolor{gray!20}{\textbf{0.00}} & 0.00 & $\text{1}^{-}$ & \cellcolor{gray!20}{\textbf{0.00}} & 0.00 & $\text{2}^{-}$ & \cellcolor{gray!20}{\textbf{0.00}} & 0.00 & $\text{1}^{-}$\\

 &  &  & 0.5 & \cellcolor{gray!20}{\textbf{35.95}} & 0.00 & \textcolor{black}{$\text{2}^{+}$} & 35.91 & 0.01 & $\text{1}^{-}$ & \cellcolor{gray!20}{\textbf{35.95}} & 0.00 & \textcolor{black}{$\text{2}^{+}$} & 35.92 & 0.00 & $\text{1}^{-}$ & 35.58 & 0.02 & $\text{2}^{-}$ & \cellcolor{gray!20}{\textbf{35.66}} & 0.01 & \textcolor{black}{$\text{1}^{+}$}\\

\multirow{-18}{*}{\centering\arraybackslash \rotatebox[origin=c]{90}{\hskip55pt scorr}} & \multirow{-10}{*}{\raggedleft\arraybackslash 10} & \multirow{-3}{*}{\raggedleft\arraybackslash 100} & 0.9 & \cellcolor{gray!20}{\textbf{36.79}} & 0.00 & \textcolor{black}{$\text{2}^{+}$} & \cellcolor{gray!20}{\textbf{36.79}} & 0.00 & $\text{1}^{-}$ & \cellcolor{gray!20}{\textbf{36.79}} & 0.00 & \textcolor{black}{$\text{2}^{+}$} & \cellcolor{gray!20}{\textbf{36.79}} & 0.00 & $\text{1}^{-}$ & 36.63 & 0.06 & $\text{2}^{-}$ & \cellcolor{gray!20}{\textbf{36.78}} & 0.00 & \textcolor{black}{$\text{1}^{+}$}\\
\cmidrule{1-22}
\cmidrule{3-22}
 &  &  & 0.1 & \cellcolor{gray!20}{\textbf{36.26}} & 0.01 & \textcolor{black}{$\text{2}^{+}$} & 36.23 & 0.01 & $\text{1}^{-}$ & \cellcolor{gray!20}{\textbf{36.29}} & 0.01 & \textcolor{black}{$\text{2}^{+}$} & 36.25 & 0.01 & $\text{1}^{-}$ & 36.08 & 0.07 & $\text{2}^{-}$ & \cellcolor{gray!20}{\textbf{36.15}} & 0.04 & \textcolor{black}{$\text{1}^{+}$}\\

 &  &  & 0.5 & \cellcolor{gray!20}{\textbf{36.75}} & 0.01 & $\text{2}^{-}$ & \cellcolor{gray!20}{\textbf{36.75}} & 0.01 & $\text{1}^{-}$ & \cellcolor{gray!20}{\textbf{36.77}} & 0.00 & \textcolor{black}{$\text{2}^{+}$} & 36.76 & 0.01 & $\text{1}^{-}$ & 36.40 & 0.11 & $\text{2}^{-}$ & \cellcolor{gray!20}{\textbf{36.74}} & 0.01 & \textcolor{black}{$\text{1}^{+}$}\\

 &  & \multirow{-3}{*}{\raggedleft\arraybackslash 25} & 0.9 & \cellcolor{gray!20}{\textbf{36.74}} & 0.01 & \textcolor{black}{$\text{2}^{+}$} & 35.67 & 0.32 & $\text{1}^{-}$ & \cellcolor{gray!20}{\textbf{36.77}} & 0.00 & \textcolor{black}{$\text{2}^{+}$} & 36.65 & 0.03 & $\text{1}^{-}$ & 36.39 & 0.04 & $\text{2}^{-}$ & \cellcolor{gray!20}{\textbf{36.75}} & 0.01 & \textcolor{black}{$\text{1}^{+}$}\\

\cmidrule{3-22}
 &  &  & 0.1 & \cellcolor{gray!20}{\textbf{36.34}} & 0.00 & \textcolor{black}{$\text{2}^{+}$} & 36.33 & 0.01 & $\text{1}^{-}$ & \cellcolor{gray!20}{\textbf{36.34}} & 0.00 & \textcolor{black}{$\text{2}^{+}$} & 36.33 & 0.00 & $\text{1}^{-}$ & \cellcolor{gray!20}{\textbf{36.00}} & 0.04 & $\text{2}^{-}$ & \cellcolor{gray!20}{\textbf{36.00}} & 0.03 & $\text{1}^{-}$\\

 &  &  & 0.5 & \cellcolor{gray!20}{\textbf{36.79}} & 0.00 & \textcolor{black}{$\text{2}^{+}$} & \cellcolor{gray!20}{\textbf{36.79}} & 0.00 & $\text{1}^{-}$ & \cellcolor{gray!20}{\textbf{36.79}} & 0.00 & \textcolor{black}{$\text{2}^{+}$} & \cellcolor{gray!20}{\textbf{36.79}} & 0.00 & $\text{1}^{-}$ & 36.61 & 0.02 & $\text{2}^{-}$ & \cellcolor{gray!20}{\textbf{36.75}} & 0.01 & \textcolor{black}{$\text{1}^{+}$}\\

 & \multirow{-10}{*}{\raggedleft\arraybackslash 2} & \multirow{-3}{*}{\raggedleft\arraybackslash 100} & 0.9 & \cellcolor{gray!20}{\textbf{36.79}} & 0.00 & \textcolor{black}{$\text{2}^{+}$} & 36.76 & 0.01 & $\text{1}^{-}$ & \cellcolor{gray!20}{\textbf{36.79}} & 0.00 & \textcolor{black}{$\text{2}^{+}$} & \cellcolor{gray!20}{\textbf{36.79}} & 0.00 & $\text{1}^{-}$ & 36.63 & 0.02 & $\text{2}^{-}$ & \cellcolor{gray!20}{\textbf{36.78}} & 0.00 & \textcolor{black}{$\text{1}^{+}$}\\

\cmidrule{2-22}
 &  &  & 0.1 & \cellcolor{gray!20}{\textbf{25.42}} & 0.16 & \textcolor{black}{$\text{2}^{+}$} & 24.34 & 0.22 & $\text{1}^{-}$ & \cellcolor{gray!20}{\textbf{25.65}} & 0.13 & \textcolor{black}{$\text{2}^{+}$} & 24.63 & 0.32 & $\text{1}^{-}$ & \cellcolor{gray!20}{\textbf{24.23}} & 0.20 & \textcolor{black}{$\text{2}^{+}$} & 22.77 & 0.39 & $\text{1}^{-}$\\

 &  &  & 0.5 & 36.68 & 0.01 & $\text{2}^{-}$ & \cellcolor{gray!20}{\textbf{36.70}} & 0.00 & \textcolor{black}{$\text{1}^{+}$} & \cellcolor{gray!20}{\textbf{36.70}} & 0.00 & \textcolor{black}{$\text{2}^{+}$} & \cellcolor{gray!20}{\textbf{36.70}} & 0.00 & $\text{1}^{-}$ & 36.02 & 0.09 & $\text{2}^{-}$ & \cellcolor{gray!20}{\textbf{36.56}} & 0.03 & \textcolor{black}{$\text{1}^{+}$}\\

 &  & \multirow{-3}{*}{\raggedleft\arraybackslash 25} & 0.9 & \cellcolor{gray!20}{\textbf{36.75}} & 0.01 & \textcolor{black}{$\text{2}^{+}$} & 36.67 & 0.06 & $\text{1}^{-}$ & \cellcolor{gray!20}{\textbf{36.77}} & 0.00 & \textcolor{black}{$\text{2}^{+}$} & 36.76 & 0.01 & $\text{1}^{-}$ & 36.15 & 0.03 & $\text{2}^{-}$ & \cellcolor{gray!20}{\textbf{36.76}} & 0.01 & \textcolor{black}{$\text{1}^{+}$}\\

\cmidrule{3-22}
 &  &  & 0.1 & \cellcolor{gray!20}{\textbf{26.03}} & 0.06 & \textcolor{black}{$\text{2}^{+}$} & 25.28 & 0.13 & $\text{1}^{-}$ & \cellcolor{gray!20}{\textbf{26.14}} & 0.03 & \textcolor{black}{$\text{2}^{+}$} & 25.47 & 0.14 & $\text{1}^{-}$ & \cellcolor{gray!20}{\textbf{0.00}} & 0.00 & $\text{2}^{-}$ & \cellcolor{gray!20}{\textbf{0.00}} & 0.00 & $\text{1}^{-}$\\

 &  &  & 0.5 & \cellcolor{gray!20}{\textbf{36.75}} & 0.00 & \textcolor{black}{$\text{2}^{+}$} & 36.74 & 0.00 & $\text{1}^{-}$ & \cellcolor{gray!20}{\textbf{36.75}} & 0.00 & \textcolor{black}{$\text{2}^{+}$} & 36.74 & 0.00 & $\text{1}^{-}$ & 35.56 & 0.03 & $\text{2}^{-}$ & \cellcolor{gray!20}{\textbf{36.10}} & 0.03 & \textcolor{black}{$\text{1}^{+}$}\\

 & \multirow{-10}{*}{\raggedleft\arraybackslash 5} & \multirow{-3}{*}{\raggedleft\arraybackslash 100} & 0.9 & \cellcolor{gray!20}{\textbf{36.79}} & 0.00 & \textcolor{black}{$\text{2}^{+}$} & \cellcolor{gray!20}{\textbf{36.79}} & 0.00 & $\text{1}^{-}$ & \cellcolor{gray!20}{\textbf{36.79}} & 0.00 & \textcolor{black}{$\text{2}^{+}$} & \cellcolor{gray!20}{\textbf{36.79}} & 0.00 & $\text{1}^{-}$ & 35.60 & 0.05 & $\text{2}^{-}$ & \cellcolor{gray!20}{\textbf{36.78}} & 0.00 & \textcolor{black}{$\text{1}^{+}$}\\

\cmidrule{2-22}
 &  &  & 0.1 & \cellcolor{gray!20}{\textbf{0.00}} & 0.00 & $\text{2}^{-}$ & \cellcolor{gray!20}{\textbf{0.00}} & 0.00 & $\text{1}^{-}$ & \cellcolor{gray!20}{\textbf{0.00}} & 0.00 & $\text{2}^{-}$ & \cellcolor{gray!20}{\textbf{0.00}} & 0.00 & $\text{1}^{-}$ & \cellcolor{gray!20}{\textbf{0.00}} & 0.00 & $\text{2}^{-}$ & \cellcolor{gray!20}{\textbf{0.00}} & 0.00 & $\text{1}^{-}$\\

 &  &  & 0.5 & \cellcolor{gray!20}{\textbf{35.15}} & 0.01 & \textcolor{black}{$\text{2}^{+}$} & 35.06 & 0.03 & $\text{1}^{-}$ & \cellcolor{gray!20}{\textbf{35.17}} & 0.01 & \textcolor{black}{$\text{2}^{+}$} & 35.08 & 0.03 & $\text{1}^{-}$ & \cellcolor{gray!20}{\textbf{35.07}} & 0.01 & \textcolor{black}{$\text{2}^{+}$} & 35.01 & 0.02 & $\text{1}^{-}$\\

 &  & \multirow{-3}{*}{\raggedleft\arraybackslash 25} & 0.9 & 36.76 & 0.01 & $\text{2}^{-}$ & \cellcolor{gray!20}{\textbf{36.78}} & 0.00 & \textcolor{black}{$\text{1}^{+}$} & \cellcolor{gray!20}{\textbf{36.77}} & 0.00 & $\text{2}^{-}$ & \cellcolor{gray!20}{\textbf{36.77}} & 0.00 & $\text{1}^{-}$ & 36.15 & 0.14 & $\text{2}^{-}$ & \cellcolor{gray!20}{\textbf{36.76}} & 0.01 & \textcolor{black}{$\text{1}^{+}$}\\

\cmidrule{3-22}
 &  &  & 0.1 & \cellcolor{gray!20}{\textbf{0.00}} & 0.00 & $\text{2}^{-}$ & \cellcolor{gray!20}{\textbf{0.00}} & 0.00 & $\text{1}^{-}$ & \cellcolor{gray!20}{\textbf{0.00}} & 0.00 & $\text{2}^{-}$ & \cellcolor{gray!20}{\textbf{0.00}} & 0.00 & $\text{1}^{-}$ & \cellcolor{gray!20}{\textbf{0.00}} & 0.00 & $\text{2}^{-}$ & \cellcolor{gray!20}{\textbf{0.00}} & 0.00 & $\text{1}^{-}$\\

 &  &  & 0.5 & \cellcolor{gray!20}{\textbf{35.22}} & 0.00 & \textcolor{black}{$\text{2}^{+}$} & 35.19 & 0.01 & $\text{1}^{-}$ & \cellcolor{gray!20}{\textbf{35.23}} & 0.00 & \textcolor{black}{$\text{2}^{+}$} & 35.19 & 0.01 & $\text{1}^{-}$ & \cellcolor{gray!20}{\textbf{35.00}} & 0.02 & \textcolor{black}{$\text{2}^{+}$} & 34.88 & 0.03 & $\text{1}^{-}$\\

\multirow{-20}{*}{\centering\arraybackslash \rotatebox[origin=c]{90}{\hskip40pt uncorr}} & \multirow{-10}{*}{\raggedleft\arraybackslash 10} & \multirow{-3}{*}{\raggedleft\arraybackslash 100} & 0.9 & \cellcolor{gray!20}{\textbf{36.79}} & 0.00 & \textcolor{black}{$\text{2}^{+}$} & \cellcolor{gray!20}{\textbf{36.79}} & 0.00 & $\text{1}^{-}$ & \cellcolor{gray!20}{\textbf{36.79}} & 0.00 & \textcolor{black}{$\text{2}^{+}$} & \cellcolor{gray!20}{\textbf{36.79}} & 0.00 & $\text{1}^{-}$ & 36.60 & 0.04 & $\text{2}^{-}$ & \cellcolor{gray!20}{\textbf{36.78}} & 0.00 & \textcolor{black}{$\text{1}^{+}$}\\
\cmidrule{1-22}
\cmidrule{3-22}
 &  &  & 0.1 & \cellcolor{gray!20}{\textbf{36.76}} & 0.01 & \textcolor{black}{$\text{2}^{+}$} & 36.74 & 0.01 & $\text{1}^{-}$ & \cellcolor{gray!20}{\textbf{36.77}} & 0.01 & \textcolor{black}{$\text{2}^{+}$} & 36.76 & 0.01 & $\text{1}^{-}$ & 36.57 & 0.07 & $\text{2}^{-}$ & \cellcolor{gray!20}{\textbf{36.74}} & 0.01 & \textcolor{black}{$\text{1}^{+}$}\\

 &  &  & 0.5 & \cellcolor{gray!20}{\textbf{36.75}} & 0.01 & \textcolor{black}{$\text{2}^{+}$} & 36.36 & 0.17 & $\text{1}^{-}$ & \cellcolor{gray!20}{\textbf{36.77}} & 0.00 & \textcolor{black}{$\text{2}^{+}$} & 36.72 & 0.02 & $\text{1}^{-}$ & 36.50 & 0.07 & $\text{2}^{-}$ & \cellcolor{gray!20}{\textbf{36.75}} & 0.01 & \textcolor{black}{$\text{1}^{+}$}\\

 &  & \multirow{-3}{*}{\raggedleft\arraybackslash 25} & 0.9 & \cellcolor{gray!20}{\textbf{36.75}} & 0.02 & \textcolor{black}{$\text{2}^{+}$} & 34.39 & 0.50 & $\text{1}^{-}$ & \cellcolor{gray!20}{\textbf{36.77}} & 0.00 & \textcolor{black}{$\text{2}^{+}$} & 36.47 & 0.15 & $\text{1}^{-}$ & 36.50 & 0.08 & $\text{2}^{-}$ & \cellcolor{gray!20}{\textbf{36.74}} & 0.00 & \textcolor{black}{$\text{1}^{+}$}\\

\cmidrule{3-22}
 &  &  & 0.1 & \cellcolor{gray!20}{\textbf{36.79}} & 0.00 & \textcolor{black}{$\text{2}^{+}$} & \cellcolor{gray!20}{\textbf{36.79}} & 0.00 & $\text{1}^{-}$ & \cellcolor{gray!20}{\textbf{36.79}} & 0.00 & \textcolor{black}{$\text{2}^{+}$} & \cellcolor{gray!20}{\textbf{36.79}} & 0.00 & $\text{1}^{-}$ & 36.68 & 0.01 & $\text{2}^{-}$ & \cellcolor{gray!20}{\textbf{36.75}} & 0.01 & \textcolor{black}{$\text{1}^{+}$}\\

 &  &  & 0.5 & \cellcolor{gray!20}{\textbf{36.79}} & 0.00 & \textcolor{black}{$\text{2}^{+}$} & \cellcolor{gray!20}{\textbf{36.79}} & 0.00 & $\text{1}^{-}$ & \cellcolor{gray!20}{\textbf{36.79}} & 0.00 & \textcolor{black}{$\text{2}^{+}$} & \cellcolor{gray!20}{\textbf{36.79}} & 0.00 & $\text{1}^{-}$ & 36.68 & 0.01 & $\text{2}^{-}$ & \cellcolor{gray!20}{\textbf{36.78}} & 0.00 & \textcolor{black}{$\text{1}^{+}$}\\

 & \multirow{-10}{*}{\raggedleft\arraybackslash 2} & \multirow{-3}{*}{\raggedleft\arraybackslash 100} & 0.9 & \cellcolor{gray!20}{\textbf{36.79}} & 0.00 & \textcolor{black}{$\text{2}^{+}$} & 36.69 & 0.04 & $\text{1}^{-}$ & \cellcolor{gray!20}{\textbf{36.79}} & 0.00 & \textcolor{black}{$\text{2}^{+}$} & \cellcolor{gray!20}{\textbf{36.79}} & 0.00 & $\text{1}^{-}$ & 36.69 & 0.01 & $\text{2}^{-}$ & \cellcolor{gray!20}{\textbf{36.78}} & 0.00 & \textcolor{black}{$\text{1}^{+}$}\\

\cmidrule{2-22}
 &  &  & 0.1 & \cellcolor{gray!20}{\textbf{29.60}} & 0.06 & \textcolor{black}{$\text{2}^{+}$} & 28.80 & 0.17 & $\text{1}^{-}$ & \cellcolor{gray!20}{\textbf{29.68}} & 0.07 & \textcolor{black}{$\text{2}^{+}$} & 29.16 & 0.13 & $\text{1}^{-}$ & \cellcolor{gray!20}{\textbf{29.08}} & 0.14 & \textcolor{black}{$\text{2}^{+}$} & 28.51 & 0.22 & $\text{1}^{-}$\\

 &  &  & 0.5 & 36.73 & 0.01 & $\text{2}^{-}$ & \cellcolor{gray!20}{\textbf{36.75}} & 0.02 & \textcolor{black}{$\text{1}^{+}$} & \cellcolor{gray!20}{\textbf{36.76}} & 0.01 & \textcolor{black}{$\text{2}^{+}$} & 36.75 & 0.01 & $\text{1}^{-}$ & 35.97 & 0.10 & $\text{2}^{-}$ & \cellcolor{gray!20}{\textbf{36.71}} & 0.00 & \textcolor{black}{$\text{1}^{+}$}\\

 &  & \multirow{-3}{*}{\raggedleft\arraybackslash 25} & 0.9 & \cellcolor{gray!20}{\textbf{36.74}} & 0.01 & \textcolor{black}{$\text{2}^{+}$} & 36.47 & 0.09 & $\text{1}^{-}$ & \cellcolor{gray!20}{\textbf{36.77}} & 0.00 & \textcolor{black}{$\text{2}^{+}$} & 36.71 & 0.03 & $\text{1}^{-}$ & 36.06 & 0.13 & $\text{2}^{-}$ & \cellcolor{gray!20}{\textbf{36.75}} & 0.01 & \textcolor{black}{$\text{1}^{+}$}\\

\cmidrule{3-22}
 &  &  & 0.1 & \cellcolor{gray!20}{\textbf{29.93}} & 0.02 & \textcolor{black}{$\text{2}^{+}$} & 29.55 & 0.14 & $\text{1}^{-}$ & \cellcolor{gray!20}{\textbf{29.96}} & 0.01 & \textcolor{black}{$\text{2}^{+}$} & 29.66 & 0.09 & $\text{1}^{-}$ & \cellcolor{gray!20}{\textbf{0.00}} & 0.00 & $\text{2}^{-}$ & \cellcolor{gray!20}{\textbf{0.00}} & 0.00 & $\text{1}^{-}$\\

 &  &  & 0.5 & \cellcolor{gray!20}{\textbf{36.79}} & 0.00 & $\text{2}^{-}$ & \cellcolor{gray!20}{\textbf{36.79}} & 0.00 & $\text{1}^{-}$ & \cellcolor{gray!20}{\textbf{36.79}} & 0.00 & $\text{2}^{-}$ & \cellcolor{gray!20}{\textbf{36.79}} & 0.00 & $\text{1}^{-}$ & 35.39 & 0.08 & $\text{2}^{-}$ & \cellcolor{gray!20}{\textbf{36.31}} & 0.03 & \textcolor{black}{$\text{1}^{+}$}\\

 & \multirow{-10}{*}{\raggedleft\arraybackslash 5} & \multirow{-3}{*}{\raggedleft\arraybackslash 100} & 0.9 & \cellcolor{gray!20}{\textbf{36.79}} & 0.00 & \textcolor{black}{$\text{2}^{+}$} & \cellcolor{gray!20}{\textbf{36.79}} & 0.00 & $\text{1}^{-}$ & \cellcolor{gray!20}{\textbf{36.79}} & 0.00 & \textcolor{black}{$\text{2}^{+}$} & \cellcolor{gray!20}{\textbf{36.79}} & 0.00 & $\text{1}^{-}$ & 35.41 & 0.04 & $\text{2}^{-}$ & \cellcolor{gray!20}{\textbf{36.78}} & 0.00 & \textcolor{black}{$\text{1}^{+}$}\\

\cmidrule{2-22}
 &  &  & 0.1 & \cellcolor{gray!20}{\textbf{0.00}} & 0.00 & $\text{2}^{-}$ & \cellcolor{gray!20}{\textbf{0.00}} & 0.00 & $\text{1}^{-}$ & \cellcolor{gray!20}{\textbf{0.00}} & 0.00 & $\text{2}^{-}$ & \cellcolor{gray!20}{\textbf{0.00}} & 0.00 & $\text{1}^{-}$ & \cellcolor{gray!20}{\textbf{0.00}} & 0.00 & $\text{2}^{-}$ & \cellcolor{gray!20}{\textbf{0.00}} & 0.00 & $\text{1}^{-}$\\

 &  &  & 0.5 & \cellcolor{gray!20}{\textbf{35.20}} & 0.02 & \textcolor{black}{$\text{2}^{+}$} & 35.13 & 0.04 & $\text{1}^{-}$ & \cellcolor{gray!20}{\textbf{35.23}} & 0.01 & \textcolor{black}{$\text{2}^{+}$} & 35.13 & 0.03 & $\text{1}^{-}$ & \cellcolor{gray!20}{\textbf{35.10}} & 0.03 & $\text{2}^{-}$ & 35.08 & 0.04 & $\text{1}^{-}$\\

 &  & \multirow{-3}{*}{\raggedleft\arraybackslash 25} & 0.9 & 36.76 & 0.01 & $\text{2}^{-}$ & \cellcolor{gray!20}{\textbf{36.77}} & 0.00 & \textcolor{black}{$\text{1}^{+}$} & \cellcolor{gray!20}{\textbf{36.77}} & 0.00 & \textcolor{black}{$\text{2}^{+}$} & \cellcolor{gray!20}{\textbf{36.77}} & 0.00 & $\text{1}^{-}$ & 36.06 & 0.08 & $\text{2}^{-}$ & \cellcolor{gray!20}{\textbf{36.75}} & 0.01 & \textcolor{black}{$\text{1}^{+}$}\\

\cmidrule{3-22}
 &  &  & 0.1 & \cellcolor{gray!20}{\textbf{0.00}} & 0.00 & $\text{2}^{-}$ & \cellcolor{gray!20}{\textbf{0.00}} & 0.00 & $\text{1}^{-}$ & \cellcolor{gray!20}{\textbf{0.00}} & 0.00 & $\text{2}^{-}$ & \cellcolor{gray!20}{\textbf{0.00}} & 0.00 & $\text{1}^{-}$ & \cellcolor{gray!20}{\textbf{0.00}} & 0.00 & $\text{2}^{-}$ & \cellcolor{gray!20}{\textbf{0.00}} & 0.00 & $\text{1}^{-}$\\

 &  &  & 0.5 & \cellcolor{gray!20}{\textbf{35.30}} & 0.01 & \textcolor{black}{$\text{2}^{+}$} & 35.26 & 0.02 & $\text{1}^{-}$ & \cellcolor{gray!20}{\textbf{35.30}} & 0.00 & \textcolor{black}{$\text{2}^{+}$} & 35.27 & 0.01 & $\text{1}^{-}$ & \cellcolor{gray!20}{\textbf{34.93}} & 0.01 & \textcolor{black}{$\text{2}^{+}$} & 34.89 & 0.02 & $\text{1}^{-}$\\

\multirow{-18}{*}{\centering\arraybackslash \rotatebox[origin=c]{90}{\hskip55pt usw}} & \multirow{-10}{*}{\raggedleft\arraybackslash 10} & \multirow{-3}{*}{\raggedleft\arraybackslash 100} & 0.9 & \cellcolor{gray!20}{\textbf{36.79}} & 0.00 & \textcolor{black}{$\text{2}^{+}$} & \cellcolor{gray!20}{\textbf{36.79}} & 0.00 & $\text{1}^{-}$ & \cellcolor{gray!20}{\textbf{36.79}} & 0.00 & \textcolor{black}{$\text{2}^{+}$} & \cellcolor{gray!20}{\textbf{36.79}} & 0.00 & $\text{1}^{-}$ & 36.53 & 0.08 & $\text{2}^{-}$ & \cellcolor{gray!20}{\textbf{36.78}} & 0.00 & \textcolor{black}{$\text{1}^{+}$}\\
\bottomrule
\end{tabular}\end{scriptsize}
\end{table*}

\section{Conclusion}
\label{sec:conclusion}

We studied the zero-one knapsack problem in the context of evolutionary diversity optimization. The goal is to evolve a set of packings which all have a minimum profit, do not violate the capacity limit of the knapsack and differ with respect to the packing structure. We presented a simple $(\mu+1)$-EA that is initialized with an approximate solution obtained by a classical FPTAS for the knapsack problem and uses entropy as a diversity measure. The algorithm is evaluated with different standard mutation operators for binary representation and two (strongly) EDO-focused mutation operators. In addition we studied the effect of an EDO-focused repair- and crossover-operator. A comprehensive study on different instances and setups showed that (a) heavy-tailed mutation is beneficial in the long-term, (b) EDO-focused operators clearly dominate standard mutation in a scenario with vigorously limited computational budget, (c) the repair operators is slightly beneficial and (d), as a negative result, the proposed crossover operator for most setups has a negative effect on the EAs' progress.

The research field of EDO is still in its infancy paving the way for future work. Certainly, studying the effects of biased mutation and crossover for other combinatorial optimization problem, e.g., vertex-cover or variants of scheduling problems, seems natural. In addition, supported by the efficient short-term progress of biased-mutation in this paper, studying EDO in the course of dynamic optimization seems promising where the instance at hand is subject to dynamic changes every $\tau$ iterations and re-optimization and -diversifying is necessary.

\section*{Acknowledgment}
This work was supported by the Australian Research Council\newline{}through grant DP190103894 and by the South Australian Government through the Research Consortium "Unlocking Complex Resources through Lean Processing".

\bibliographystyle{ACM-Reference-Format}
\bibliography{arxiv}

\appendix

\end{document}